\documentclass[letterpaper]{article} 
\usepackage{aaai2026}  
\usepackage{times}  
\usepackage{helvet}  
\usepackage{courier}  
\usepackage[hyphens]{url}  
\usepackage{graphicx} 
\usepackage{natbib}  
\usepackage{caption} 
\usepackage{algorithm}

\usepackage{newfloat}
\usepackage{listings}
\DeclareCaptionStyle{ruled}{labelfont=normalfont,labelsep=colon,strut=off} 
\floatstyle{ruled}
\newfloat{listing}{tb}{lst}{}
\floatname{listing}{Listing}
\usepackage{algpseudocode}
\usepackage{subcaption} 
\usepackage{booktabs}
\usepackage{xcolor} 
\usepackage[colorinlistoftodos,bordercolor=orange,backgroundcolor=orange!20,linecolor=orange,textsize=scriptsize]{todonotes}
\usepackage{soul}
\usepackage{enumitem} 
\nocopyright

\title{DeToNATION: Decoupled Torch Network-Aware Training~on~Interlinked~Online~Nodes} 
\author{
    Mogens Henrik From,
    Jacob Nielsen,
    Lukas Galke Poech,
    Peter Schneider-Kamp
}
\affiliations{
	Department of Mathematics and Computer Science, University of Southern Denmark, Odense, Denmark\\
	from@imada.sdu.dk, jacn@imada.sdu.dk, galke@imada.sdu.dk, petersk@imada.sdu.dk
}

\begin{document}

\maketitle

\begin{abstract}
Training large neural network models requires extensive computational resources, often distributed across several nodes and accelerators. Recent findings suggest that it may be sufficient to only exchange the fast-moving components of the gradients, while accumulating momentum locally (Decoupled Momentum, or DeMo). However, DeMo assumes that models fit on a single accelerator. We relax this assumption and introduce FlexDeMo, whereby nodes fully shard model parameters locally across different accelerators, while inter-node communication is reduced by synchronizing only fast-moving components instead of the full gradients -- resulting in a hybrid sharded data parallel training strategy. We further introduce a framework, called DeToNATION, that generalizes DeMo, FlexDeMo, and other popular distributed training schemes such as DiLoCo -- introducing new variations of replication schemes and challenging choices made in DeMo. Our results across language and vision domains show that FlexDeMo attains similar validation loss to hybrid sharded data parallel training employing AdamW and full gradient synchronization, while being substantially faster. FlexDeMo is thus a promising distributed training scheme for the largest machine learning models.
\end{abstract}

\begin{links}
    \link{Code}{github.com/schneiderkamplab/DeToNATION/}
\end{links}

\section{Introduction}
Training large deep neural networks (DNNs) induces large amounts of network traffic in the form of gradients that are transmitted between accelerators, typically requiring expensive localized high-throughput network setups on high-performance computing clusters. The network throughput increasingly becomes a bottleneck, as the number of accelerator nodes participating in the training increases and the general network congestion increases. Recent work shows that synchronizing the full optimizer state is not always necessary for state-of-the-art results through decoupling momentum updates and allowing controlled divergence within each rank by carefully controlled inter-accelerator communication termed Decoupled Momentum optimization (DeMo)~\cite{peng2024demodecoupledmomentumoptimization}.

DeMo presents a viable strategy for distributed training with reduced gradient communication and, thus, enables distributed data parallel (DDP) training of DNNs with relatively low network bandwidth requirements. However, this strategy comes with several caveats. First, relying on DDP implies the constraint that the DNN model and optimizer states must fit within the memory capacity of each accelerator. This severely limits the applicability for training large models such as state-of-the-art large language models, which commonly do not fit within the memory of a single accelerator. Second, DeMo inherently relies on a distributed gathering operation whose bandwidth requirements scale linearly with the number of accelerators. Third, DeMo leaves many open questions regarding the replication scheme and hyperparameters, such as chunk size, TopK and employing the sign-function.

In this work, we introduce the Flexible Decoupled Momentum optimizer (FlexDeMo) for combining fully sharded data parallel (FSDP) training with decoupled momentum updates. This optimizer employs a hybrid-sharding strategy, where the model and optimizer states are typically sharded intra-node and replicated between nodes.

Instead of synchronizing the full gradients between the nodes as in extant hybrid sharding strategies, we compress, gather, and decompress selected relevant fast-moving momentum components following the approach introduced by DeMo for DDP training~\cite{peng2024demodecoupledmomentumoptimization}. This relaxes constraints regarding the accelerator memory requirements while simultaneously reducing inter-node bandwidth requirements. 
We further investigate hyperparameters of DeMo and challenge the fast-moving momenta-based replication scheme by introducing Striding, Random and the previously proposed DiLoCo schemes. This addresses the three caveats of DeMo mentioned above. First, FlexDeMo allows for decoupled momentum training of models that do not fit into the memory of a single accelerator but fit into the combined memory of the accelerators of one node. Second, FlexDeMo replicates between nodes rather than accelerators, effectively reducing the bandwidth requirements of the distributed gathering operation, which now scales linearly with the number of nodes rather than the number of accelerators. Third, we justify our choice of hyperparameters by conducting experiments.

We validate our results on the T5, ViT, and OLMo models in machine translation, image classification and causal language modeling, respectively. 
We show that FlexDeMo is faster than both DeMo and extant hybrid sharding strategies while achieving comparable validation loss.
In essence, FlexDeMo enables training of larger DNNs more efficiently across multiple nodes of a cluster or even across geographically-dispersed clusters. We show that the Random replication scheme, given the same bandwidth usage significantly outperforms DeMo and other replications schemes in translation, whereas we show DeMo to be superior in image-classification and causal language modeling. Finally, we show that sharing the signed gradients is clearly beneficial.

In summary, our contributions are as follows:
\begin{itemize}[nosep]
	\item The first implementation of a hybrid sharded training strategy combining intra-node FSDP with decoupled momentum optimization (FlexDeMo). 
	\item Increased efficiency in bandwidth-limited settings, both compared to FSDP with full gradient synchronization and previous work (DeMo, DiLoCo).
	\item Comparable model performance compared to hybrid sharded training strategies using the AdamW optimizer. 
	\item Introducing Random, Striding and DiLoCo replication schemes into a new DeToNATION framework.
	\item Introducing a decoupled variant of AdamW.
	\item An analysis of the critical hyper parameters TopK, chunk size and sign providing guidance on their effects on efficiency and model performance.
\end{itemize}

\section{Related Work}
Strategies to both scale and accelerate the training of deep neural networks have been active research areas \cite{shoeybi2019megatron,rajbhandari2020zero} for several years. The importance of these methods' effectiveness is increasing as we train more and more large language models both in and across growing HPC capacities. We will first clarify the terminology and then provide a brief overview of the most important advances in distributed training.

\emph{Distributed data parallel (DDP)} replicates the model and optimizer states across multiple individual processes, which each handle a subset of the training data. The gradients are averaged from all processes to keep the model weights synchronized.  
In \emph{model parallelism}, the model is split across accelerators, consequently increasing the communication overhead as each device computes only the forward and backward passes for its assigned model-parts, requiring the communication of the intermediate activation between devices.
\emph{Tensor parallelism} is similar to model parallelism but even individual tensors are split across devices. 

The zero redundancy optimizer (ZeRO)~\cite{rajbhandari2020zero}, tackling data and model parallelism, has introduced a variety of strategies to train large models, including partitioning optimizer states, gradients, and parameters -- all of which are pulled dynamically to a single accelerator on demand. ZeRO has been integrated in the DeepSpeed library~\cite{rasley2020deepspeed}. The DeepSpeed team has subsequently extended ZeRO with releases of Zero-2~\cite{zero2} and Zero-3~\cite{zero3}.
While ZeRO aims to achieve memory efficiency to enable training of very large models, it also comes with a substantial communication overhead. Both DeMo~\cite{peng2024demodecoupledmomentumoptimization} and our proposed FlexDeMo alleviate this communication overhead.

FSDP~\cite{zhao2023fsdp} takes inspiration from ZeRO but modifies how gradients are exchanged: ZeRO uses a reduce-scatter operation to distribute the gradients and an all-gather operation for the updated parameters. FSDP instead uses an all-gather operation to recover unsharded parameters and a reduce-scatter operation for the gradients.
FSDP also introduces a \emph{hybrid sharding} strategy, enabling users to define sharding groups to trade-off memory and throughput.

Another common strategy is to optimize locally for multiple steps before synchronization~\cite{DBLP:conf/iclr/Stich19}.
DiLoCo~\cite{douillard2023diloco} employs federated averaging to enable training in poorly connected nodes through parallel local optimization and periodic global averaging. 

Notably, there are several other advances in accelerating large-scale neural network training, such as gradient compression by only considering its sign~\cite{bernstein2018signsgd,DBLP:conf/nips/JiangYYZ24} or through low-rank projection of the gradients~\cite{DBLP:conf/icml/Zhao0CWAT24}. Although these approaches tackle similar challenges, we consider these directions orthogonal to our work, as we are interested in reducing the communication overhead regarding optimizer states.

\section{Methods}\label{sec:method}
We first present our hybrid-sharded decoupled momentum optimizer, FlexDeMo, which allows divergent optimizer states across accelerator nodes. Sharding to the accelerators of a node and replicating between the nodes is a reasonable strategy. We introduce the Random and Striding replication schemes, compare with the previously proposed DiLoCo, and challenge DeMo's effectiveness both regarding convergence and computation. 
FlexDeMo can be used to shard and replicate between any sets of accelerators, though, allowing both replicating model shards to multiple subsets of accelerators on one node and distributing different shards across multiple nodes. Thereby, we are relaxing the model memory constraint by sharding the model and optimizer states across multiple accelerators, typically within one node, it becomes possible to train significantly larger models. DeMo replication is guided by the fast moving components of the momentum, computed via the discrete cosine transform (DCT-II), $n$-stride indices for Striding and, naturally, randomly selected for Random.

One of the main practical limitations of DeMo is that it does not support FSDP or other sharding strategies at all and relies on an \texttt{all\_gather} operation across multiple nodes, which is inherently slow, as it requires every training process to exchange and receive gradients from every other instance (see Appendix A).
This raises the new research question regarding whether DeMo would be suitable in a sharded setting such as FSDP, as well as investigating if and how its hyperparameters need to be adjusted -- in particular, how many fast-moving components to extract and further, if it is optimal to synchronize such components. 
Here, we overcome these technical hurdles and investigate and study efficiency characteristics of such distributed training mechanism.

\subsection{FlexDeMo}
During training, the optimizer states are communicated between the accelerators within the node. Then fast moving components are extracted following the method from DeMo~\cite{peng2024demodecoupledmomentumoptimization}. These components are compressed and exchanged between groups of accelerators. This method allows for training larger models, keeping the expensive communication within the groups of accelerators, and minimizing the expensive communication between groups.

The flexible decoupled momentum training strategy we propose can be powered by multiple replication schemes; The DeMo based Discrete Cosine Transform (DCT) of the momenta, Random, Striding and DiLoCo.
The replication of the optimizer states is done between the groups of accelerators instead of between all accelerators and by only communicating the selected components (gradients). That is, accelerator 0 of node 0 replicates momentum components to accelerator 0 of node 1, and so on (detail in Appendix A). With this method, there is no replication of data between accelerator 0 in one node and accelerator 1 in another, drastically limiting the amount of data shared between nodes.

While we only share the selected gradients (e.g.: fast moving momentum components) between the nodes, and even only between accelerators with corresponding shards, we operate on all gradients within each group, where the transfer speeds are usually significantly higher. This takes advantage of the typically high bandwidth within nodes, while acknowledging the lower bandwidth between nodes.

To achieve this, we reimplement the stochastic gradient descent (SGD) with momentum optimizer from \cite{peng2024demodecoupledmomentumoptimization} (denoted DeMo-SGD), introducing a series of changes to support decoupled optimization in FSDP employing intra-node hybrid sharding. We assume that the model is wrapped as an FSDP object. Both the forward and backward passes must be wrapped in a \texttt{no\_sync} context manager, disabling automatic synchronization of gradients $\theta_t$, affecting the default \texttt{all\_reduce}-functionality, consequently decoupling the momentum $m$ across accelerator-nodes. PyTorch's \texttt{Autograd} produces the gradient for the whole unsharded grad-parameter accessible in \texttt{p.grad}.

We employ the reduce-scatter operation, averaging and then sharding the computed gradients back to their respective ranks (hence sharded parameter-size) in the sharding-parallel-group. This allows us to work only on the shards in the subsequent operations. 
We denote the two communication-groups as $S$ and $R$ referring to the sharding-group and replication group, respectively. Gradients are scattered intra-node locally within $S$ and the fast components, $q$, are communicated inter-node in $R$. We describe the operation of FlexDeMo in Algorithm~\ref{alg:demo} and provide a diagram of the communication in Appendix A.

A major difference to pure DeMo, is that it does not employ any sharding, and is thus using an \texttt{all\_gather} operation across the nodes, in the case of multiple nodes. This high amount of inter-node communication, is clearly evident in the figure in Appendix A.

\renewcommand{\algorithmicrequire}{\textbf{Input:}}
\renewcommand{\algorithmicensure}{\textbf{Output:}}
\definecolor{myblue}{HTML}{27547b}
\definecolor{mygreen}{HTML}{45b778}

\begin{algorithm*}
	\caption{\textcolor{mygreen}{FlexDeMo} extended from DeMo}
	\begin{algorithmic}
		\Require learning rate $\eta$, decay $\beta \in (0,1)$, parameters $\theta_t$, momentum $m_t$, 
		\Require \textcolor{mygreen}{sharding-set $S$, replication-set $R$,} hyperparameters $s, k$
		\State \textcolor{mygreen}{$\theta_t^i \leftarrow GradReduceScatter(\theta_t, S)$ \Comment{Get local parameter shard. Intra-Node.}}
		\State $\Delta_t^i \leftarrow LocalSGD(\theta_t^i )$ \Comment{Get local gradient $\Delta_t$}
		\State $m_t \leftarrow \beta m_t + \Delta_t$ \Comment{Accumulate gradient in momentum $m$}
		\State $q_t \leftarrow  ExtractFastComponents(m_t, s, k)$  \Comment{Extract fast components $q$ from $m$}
		\State $m_{t+1} \leftarrow m_t - q_t$ \Comment{Remove q from m}
		\State \st{$Q_t^i \leftarrow Synchronize(q_t)$} \textcolor{mygreen}{$Q_t^i \leftarrow Synchronize(q_t, R)$} \Comment{\textcolor{mygreen}{Synchronize across all nodes. Inter-Node.}}
		\State $\theta_{t+1}^i \leftarrow \theta_t^i - \eta Q_t$ \Comment{Parameter update step}
	\end{algorithmic}
	\label{alg:demo}
\end{algorithm*}

FlexDeMo degrades gracefully to pure FSDP and pure DDP settings for trivial sharding and replication groups: In the edge-case of only sharding ($\left|R\right|$ = 1), the behaviour of FlexDeMo corresponds to FSDP. In the edge-case of only replication ($\left|S\right|$ = 1), the behavior of FlexDeMo corresponds to DDP with DeMo-style replication. Without sharding and replication, the behavior of FlexDeMo collapses to single-accelerator training with the underlying optimizer.

\subsection{Replication Schemes}
We are introducing the Random and Striding replication schemes and compare with the already proposed DiLoCo. Random replication denotes a random selection of $n$ indices. Striding denotes the selection of every $n$'th index, and DiLoCo denotes synchronization every n'th optimization step. These replicators do not operate on chunks of the gradients as the DeMo replicator. All replication schemes can achieve the same relative data compression. However, on top of the limited amount of gradient values shared, Striding and Random schemes do not need to share the selected indices of the gradients, as they can be reproduced across training-instances with a fixed seed, effectively halving the data transfer, for the same amount of shared gradients.

\subsection{Decoupled AdamW}\label{sec:decoupled_adamw}
AdamW is defacto the default optimizer for most objectives, prompting us to believe that it would demonstrate strong performance with these replication schemes. We implement AdamW without synchronizing the first and second momenta, which would require communicating 2-3 times more data. 
The exponential moving average (EMA) and the moving average of the squared gradients are therefore not synchronized, and neither are the max of all exponential moving averages of squared gradients, if \texttt{amsmax} is employed.

\section{Experiments}\label{sec:foundational}
We present a series of experiments on FlexDeMo covering encoder-decoder language models, vision transformers, and decoder-only language models including scaling experiments, demonstrating the capabilties and training behavior across domains.
We compare our Decoupled AdamW to DeMo-SGD as the underlying optimizer, challenging the DeMo replication scheme with our introduced Striding and Random schemes, and further, compare with DiLoCo. Last, we demonstrate the performance and efficiency comparing performance to standard Hybrid-FSDP with a standard AdamW optimizer~\cite{kingma2014adam, adamw} demonstrating the effectiveness of FlexDeMo. 

\paragraph{Overall Setup}
First, we study hyperparameters for DeMo replication, including \texttt{sign}, TopK, transfer-\texttt{dtype}. Experiments can be found in Appendix B. Summarized, all replicators show that only sharing the sign of the gradients is beneficial for the loss convergence. It is also clear that \textit{fp32} converges faster than \textit{fp16} for the dtype, again for all replication schemes. As for TopK and chunk size, the conclusion is less clear. For some cases, larger chunk sizes converge faster, while for other the optimum is lower chunk sizes. For all experiments conducted, we choose $\mathrm{chunksize}= 32$, dtype=\textit{fp32}, and syncing only the sign of the gradients. TopK is varied through the experiments, to control the compression ratio, given by $\frac{\mathrm{topK}}{\mathrm{chunksize}}$. The compression rate is the percentage of gradients shared between nodes.

In the upcoming experiments we will demonstrate the performance across sequence-to-sequence, image-classification and causal language modeling, using the T5-base, ViT-B and OLMo2 models. The experiments are carried out on a HPC with a dragonfly network architecture with 200 Gbps interconnect. Nodes are NUMA-nodes and comprised of two GPU modules each with 4 AMD MI250x GPUs (128GB per module). The two modules are connected with infinity fabric connections (50+50 Gb/s) through which they communicate.

\subsection{T5: Translation Task}\label{sec:t5:translation}
\paragraph{Setup} We train a T5-base model on the French-to-English subset of Opus Books dataset~\cite{tiedemann-2012-parallel}. We employ the SGD optimizer at a learning rate of $10^{-3}$, with Random, DeMo, DiLoCo and Striding replication. We hold out 20\% as validation data.

\paragraph{Results}
Figure \ref{fig:random:t5:valid} shows the validation loss. Training loss is available in Appendix C.
Random replication performs best, with compressions $\frac{1}{2}$ and $\frac{1}{4}$ being best, respectively. DeMo $\frac{1}{8}$ comes in second followed by DeMo $\frac{1}{4}$, $\frac{1}{16}$ and $\frac{1}{32}$. 
DiLoCo and Striding replication schemes generally converge much slower. We see similar results for Random $\frac{1}{16}$ and $\frac{1}{32}$, where the signal seems to become too sparse.

\begin{figure}[htb]
	\centering
	\includegraphics[width=\linewidth]{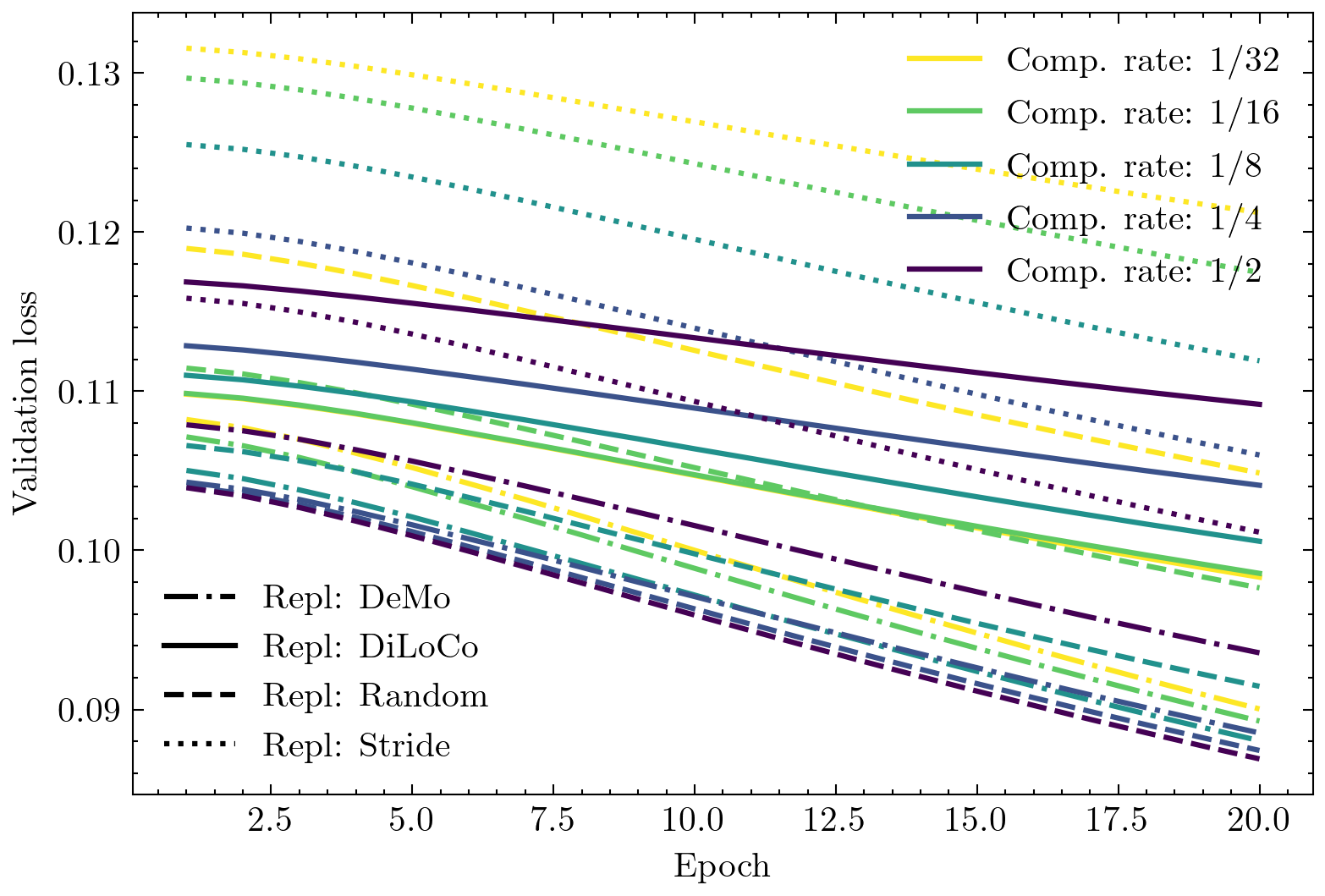}
	\caption{T5-Large Validation loss on the Opus Books En-Fr subset. Random and DeMo replication demonstrates strong performance.}
	\label{fig:random:t5:valid}
\end{figure}
\begin{figure}[t]
	\centering
	\includegraphics[width=\linewidth]{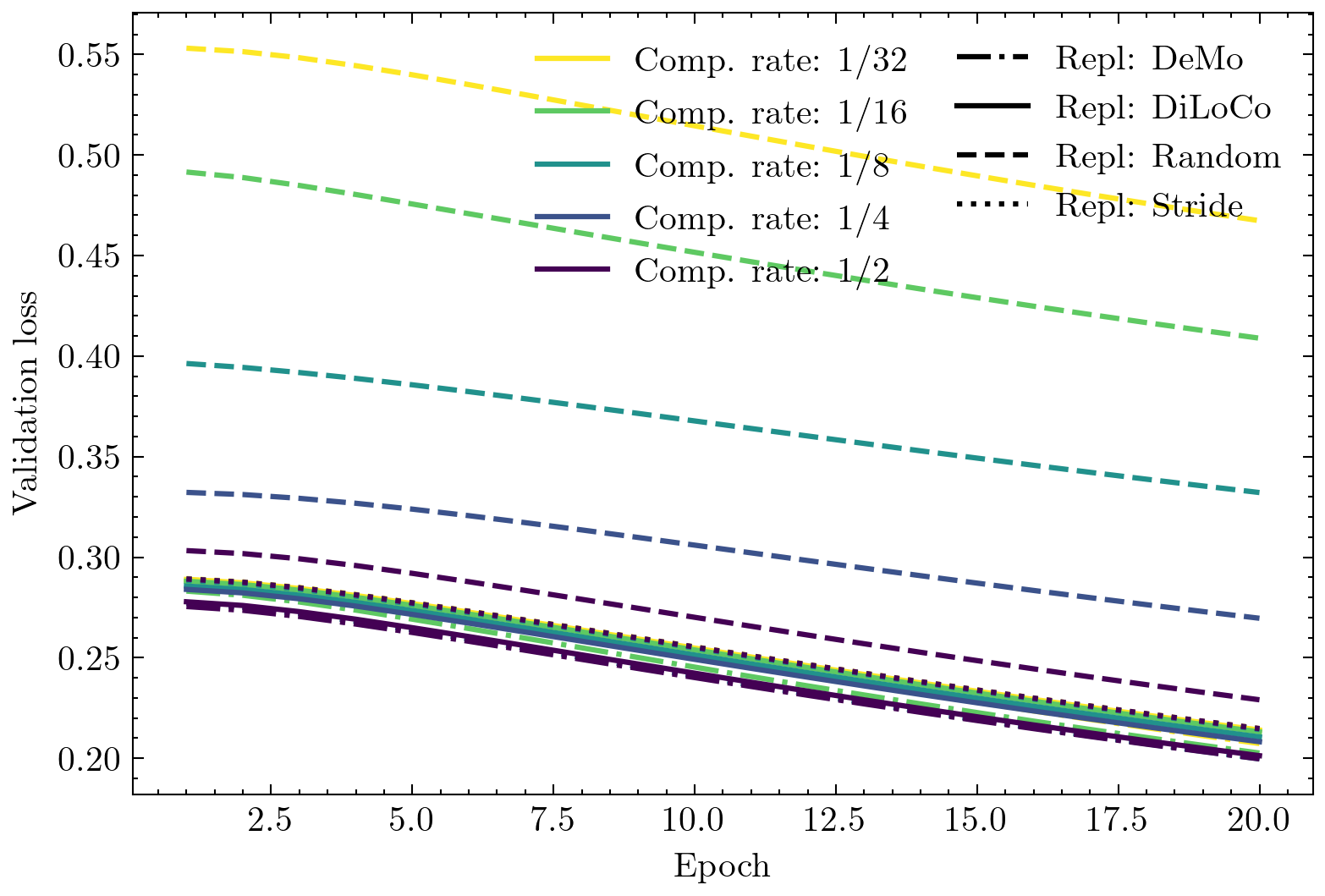}
	\caption{ViT-B Validation loss on Cifar100. DeMo and DiLoCo perform  best and similar to each other.}
	\label{fig:random:cifar100:valid}
\end{figure}

\paragraph{Discussion}
We have shown that DeMo-SGD with Random-replication schemes is best for T5 of the replication schemes, with compression rates $\frac{1}{2}$ and $\frac{1}{4}$, whereas DeMo takes second place, with compression rates $\frac{1}{8}$ and $\frac{1}{4}$, respectively. While DeMo enables more compression in its best-performing configurations, Random yields lower loss and is faster in practice. Striding and DiLoCo are converging slowly, making them less competitive here. 

\subsection{ViT: Image Classification}\label{sec:vit:imageclassification}
\paragraph{Setup}
We train a ViT-B (Patch $16$, $224$x$224$) model for vision classification on the Cifar100 dataset~\citep{krizhevsky2009learning}. Experiments are run with DeMo-SGD as the underlying optimizer with a learning rate of $10^{-5}$.

\paragraph{Results} We report our findings in Figure~\ref{fig:random:cifar100:valid}. Training loss is available in the Appendix D. 
DeMo replication with compression rates $\frac{1}{2}$ and $\frac{1}{4}$ demonstrates the highest performance (comparable), closely followed by DeMo at $\frac{1}{16}$, with DiLoCo $\frac{1}{2}$ in-between the two. We generally see that the Random scheme struggles in this domain. Even Striding replication performs better than Random.  

\paragraph{Discussion}
Contrary to the encoder-decoder architecture on a next-token-prediction training with superior Random replication scheme, DeMo replication is superior here. We hypothesis that this is because fast-moving momenta is more suited for this task, as less pronounced features do not contribute to the learning-objective, potentially inducing noise, whereas these small nuances seem to be valuable for language modeling. Randomly chosen indices can include feature-information not important for the implicit supervised contrastive learning. DiLoCo with a low compression follows DeMo's performance closely, while taking a hit on the performance when increasing the compression rate. Striding does follow DiLoCo's performance closely, we hypothesis that the structure provided by this scheme work in highly structured data, that images are.

\subsection{OLMo2: Causal Language Modeling}\label{sec:olmo:casuallm}
\paragraph{Setup}
We train the OLMo2 decoder-only model for causal language modeling, in a Hybrid-FSDP setup with $2$ nodes with $4$ accelerators each for 1B parameters. We train on the Dolma v1.6 dataset~\cite{dolma} for 10K steps. We use the OLMo's standard parameters\footnote{https://github.com/allenai/OLMo/blob/main/configs/official-0425/OLMo2-1B-stage1.yaml}, with the chance of 4\% warm-up steps. 
Each node is composed by a custom Ampere A100 64GB GPU, with 2x dual-port HDR network interface (400Gbps aggregated). The nodes are connected in a InfiniBand-based Dragonfly+ topology with a low latency interconnect with 200Gb/s.

\begin{figure}[htb]
	\centering
	\includegraphics[width=0.95\linewidth]{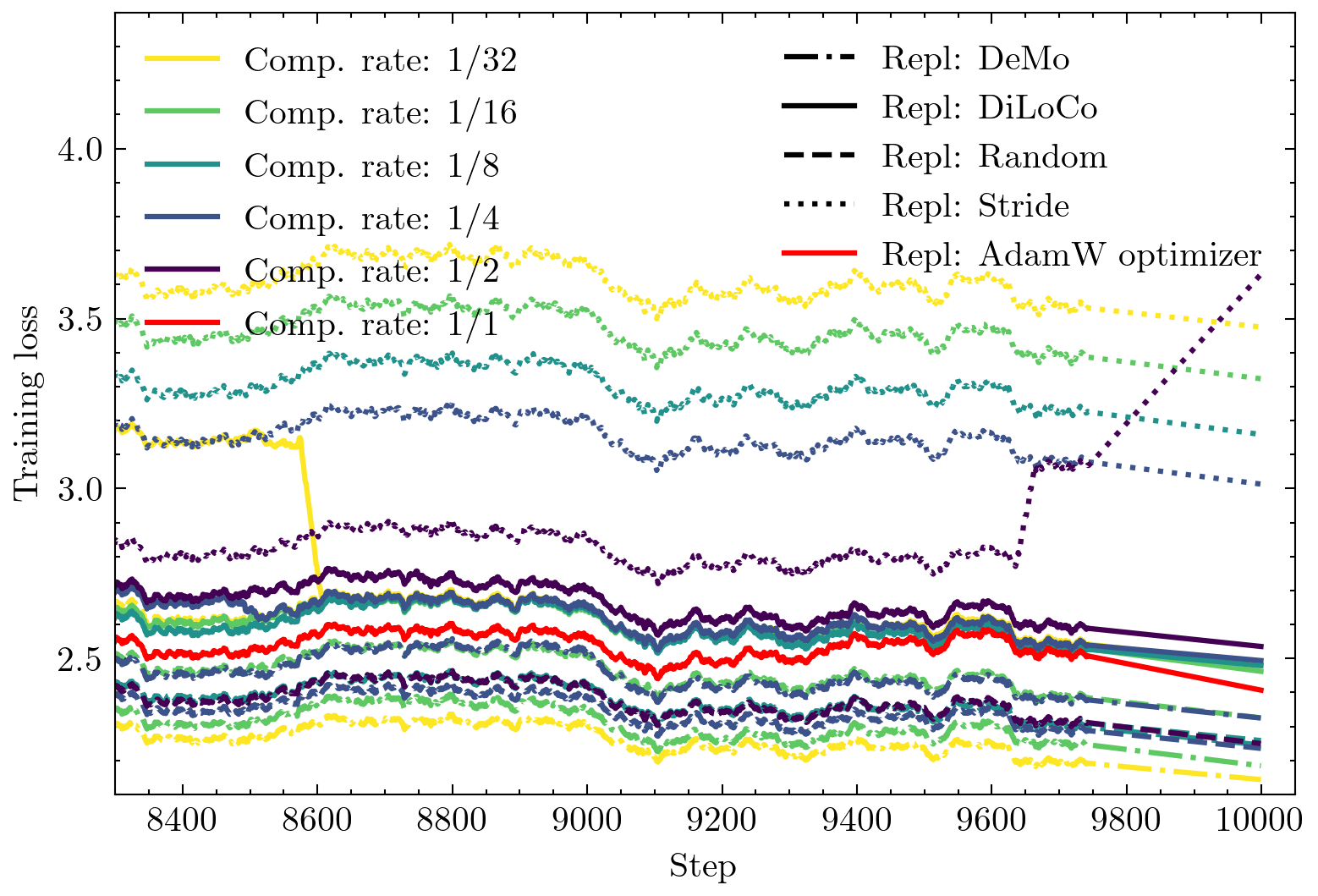}
	\caption{OLMo2 1B train loss (zoomed) over 10K training steps on Dolma v1.6 using different replicators and compression rates. All experiments, except the Hybrid-FSDP baseline with AdamW on two nodes, use DeMo-SGD.}
	\label{fig:olmo:train-loss}
\end{figure}

\begin{figure}[htbp]
	\centering
	\includegraphics[width=0.95\linewidth]{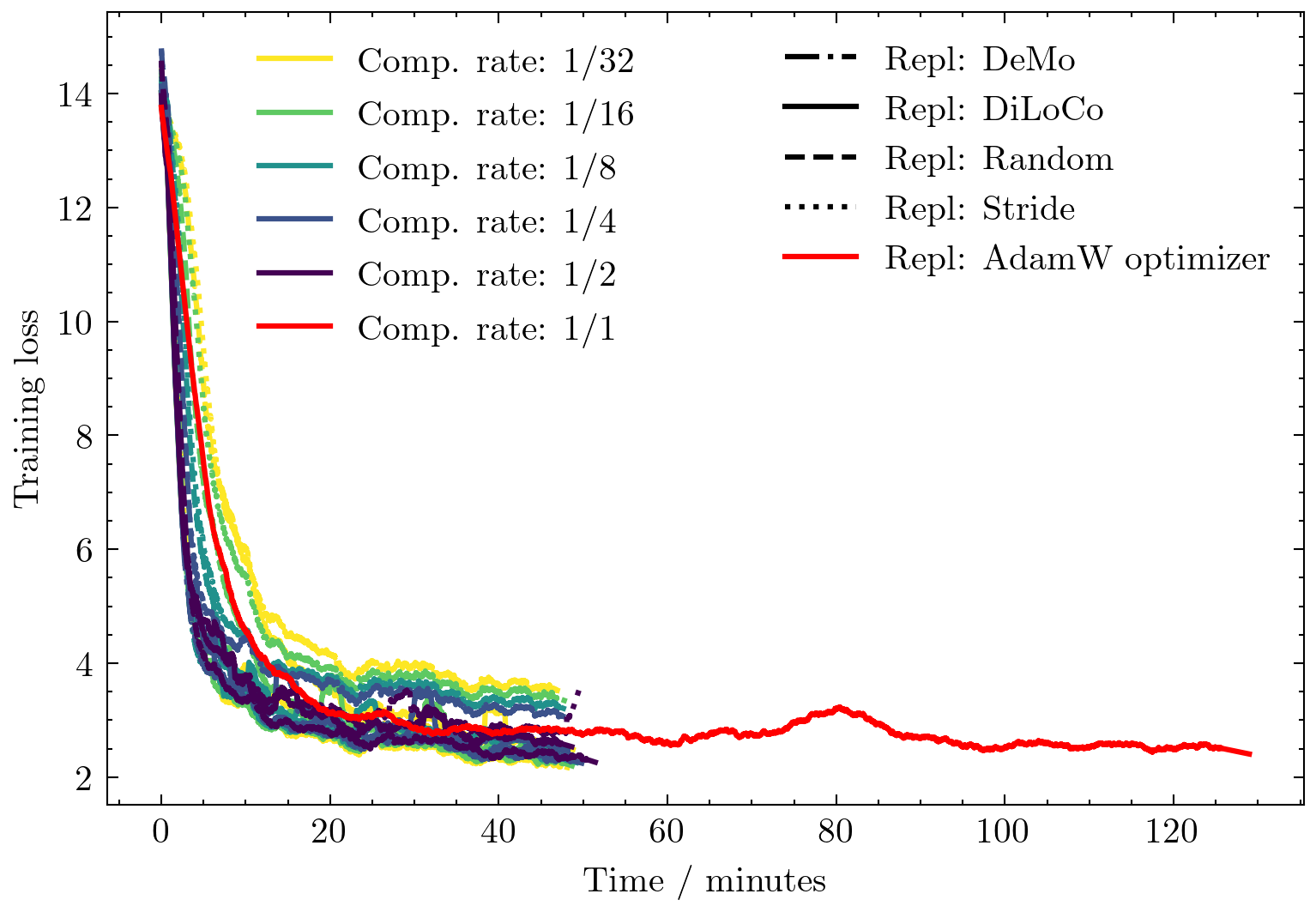}
	\caption{OLMo2 1B train loss vs. wall-clock time over 10K training steps on Dolma v1.6, comparing different replicators and compression rates. All experiments use DeMo-SGD except for the Hybrid-FSDP baseline with AdamW. FlexDeMo shows clear improvements in convergence speed.}
	\label{fig:random:train-loss-time}
\end{figure}

\paragraph{Results} We report our findings in Figures \ref{fig:olmo:train-loss} and \ref{fig:random:train-loss-time}. All of the FlexDeMo with DeMo replication demonstrates the best performance, with interestingly the compression rate $\frac{1}{32}$ performing best, closely followed by $\frac{1}{16}$, displaying some interesting artifacts of high compression. Random-replication is second-best, and better than DeMo with compression $\frac{1}{4}$, which is comparable to Random $\frac{1}{16}$. Hybrid-FSDP with conventional AdamW (red), performs better than all DiLoCo configurations, but worse than all DeMo and Random configurations. Strided replication generally yields unstable training and uncompetitive results. In Figure \ref{fig:random:train-loss-time}, we see that all replicator-types across all compression rates are substantially faster than conventional AdamW, which employs full replication. Conducting wall-time measurements on an HPC is not perfectly reliable, as they can be affected by the network-congestion during the time of the experiment, preventing exact comparisons between replicators. However, the measurements still provide insights in the general tendencies, and scale of expected runtime.

\paragraph{Discussion}
Similarly as for the ViT architecture the DeMo replicator performs best for OLMo architecture with compression rates $\frac{1}{32}$ and $\frac{1}{16}$, being the best. This is somewhat closely followed the Random replication scheme, however with larger compressions on $\frac{1}{4}, \frac{1}{8}$ and $\frac{1}{2}$, respectively, making them even less attractive to the DeMo variant. Interstingly, we do not see an optimal replication scheme within each 
We compared all FlexDeMo replicators to Hybrid-FSDP with AdamW on two HPC nodes. We see that all the replicator schemes yields training times around 2.6 times faster than the Hybrid-FSDP with AdamW setting. As timing on HPC systems is dependent on the current network congestion, we do not comment on smaller differences between the replicators. However, this does not make the large relative margin to the conventional setup uncertain, and paves the way for better utilizing GPU allocations and lowering the environmental impact substantially. We can thus see that it is clearly beneficial that the \texttt{all\_gather} is done only once per node with the hybrid-sharding strategy, instead of once per accelerator in the original decoupled momentum implementation.
With hybrid-sharding, we still need to use the \texttt{gather}-operation intra-node, which is reasonably fast, but can be optimized by implementing hooks and streams. We leave this for future work.

\subsection{Replicator bandwidth usage}\label{sec:bw_use}
\paragraph{Setup} Here we measure the actual bandwidth usage for the DeMo replicator, Random replicator and full replication, while training a T5-small model with the French-English subset of the OpusBooks dataset for 20 epochs. The setup is based on 8 RTX A6000 GPUs, in two nodes of 4 cards each, connected together with a 10 Gbps network interface. The environment is entirely controlled, with no other users, to allow truthful bandwidth measurements. The replicators are running at a compression rate of $1/16$

\paragraph{Results and Discussion} We measured an average bandwidth usage of $1070$ Mbps for full replication, $291$ Mbps for DeMo, and $152$ Mbps for Random. That is, DeMo uses about twice the bandwidth as Random, and full replication uses 7 times more. Full replication uses close to the absolute limit of the cable, and we expect that it is indeed bottlenecked by the bandwidth, explaining why it doesn't reach 16 times the use of the other replicators. We expect that DeMo use twice the bandwidth of Random, as DeMo is sharing both the gradient values, and the index of the shared gradients, due to the nature of the compression scheme. Random avoids this, by seeding the random generator on all nodes, so the indices are calculated locally, and thus not transferred. Consequently, Random can effectively share twice the amount of gradients for the same bandwidth cost.

\subsection{Performance in congested networks}
\paragraph{Setup} We use a 2 node setup, each with 4 RTX A6000 GPUs, and connected with 10 Gbps ethernet, to train T5-Large and ViT-B at different bandwidth limits. As most HPCs are subject to congested networks when training at scale, we simulate such a limited network, by limiting the bandwidth of the inter-node connection. Here, we impose the limit using Linux Traffic Control on the nodes. We analyze the average time per training step for $10$, $100$, $1{,}000$ and $10{,}000$ Mbps connections, and compare DeMo and Random replication, with underlying DeMo-SGD optimizer, with compression rates $\frac{1}{16}$ and $\frac{1}{32}$, against the Decoupled-AdamW optimizer (ours). 

\paragraph{Results} We report our results for ViT in Figure \ref{fig:avgstepspeed}, and for T5 in Appendix D. As expected, as the bandwidth decreases, the importance of the compression rate becomes more pronounced, where the compression of $\frac{1}{32}$ clearly demonstrates to be fastest followed by $\frac{1}{16}$ with $\frac{1}{1}$, being last, dictated by the amount of data to communicate, with Random DeMo-SGD being the fastest, as shown both in Figure \ref{fig:avgstepspeed:t5} and \ref{fig:avgstepspeed:vit}. This is crucial at bandwidths lower than 500Mbps, a level of bandwidths comparable to those available in HPCs in practice. Random replication is approximately $3.33$ times faster than DeMo-replication at $10$Mbps, while being approx. $18$ times faster than Decoupled-AdamW with full replication (Figure~\ref{fig:avgstepspeed:vit}).
Generally we see that Random replication with compression $\frac{1}{16}$ scales similarly to DeMo replication with compression $\frac{1}{32}$, demonstrating exactly that Random replication is around twice as fast as DeMo, as expected from DeMo transferring twice the amount of data, at the same compression rate. 

\begin{figure}[htb]
	\centering
	\includegraphics[width=\linewidth]{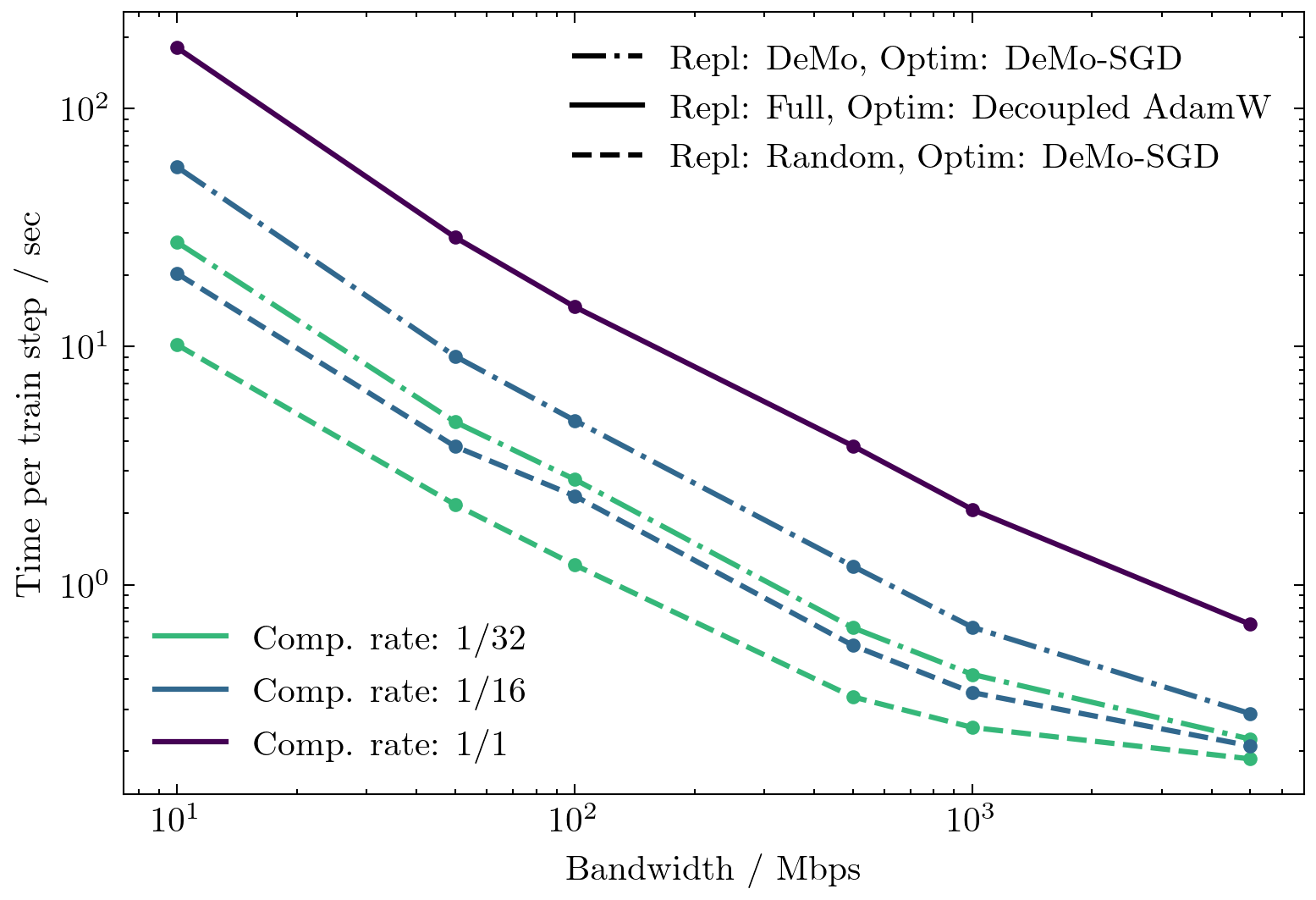}
	\caption{Average time per optimizer step for ViT-B on two nodes across employing different replicators using DeMo-SGD and Decoupled AdamW as base optimizers}
	\label{fig:avgstepspeed:vit}
	\label{fig:avgstepspeed}
\end{figure}

\subsection{OLMo2: Scaling Up}
\paragraph{Setup} In this experiment we investigate the performance and efficiency of FlexDeMo at scale. Our setup is based on the same experiments as the previous OLMo2 experiment, using the Demo-SGD optimizer, and both DeMo and Random-replication schemes with a $\frac{1}{32}$ compression rate and a learning rate of $10^{-3}$. We scale to 64 nodes \`a 4 GPUs.

\begin{figure}[htb]
	\centering
	\includegraphics[width=0.95\linewidth]{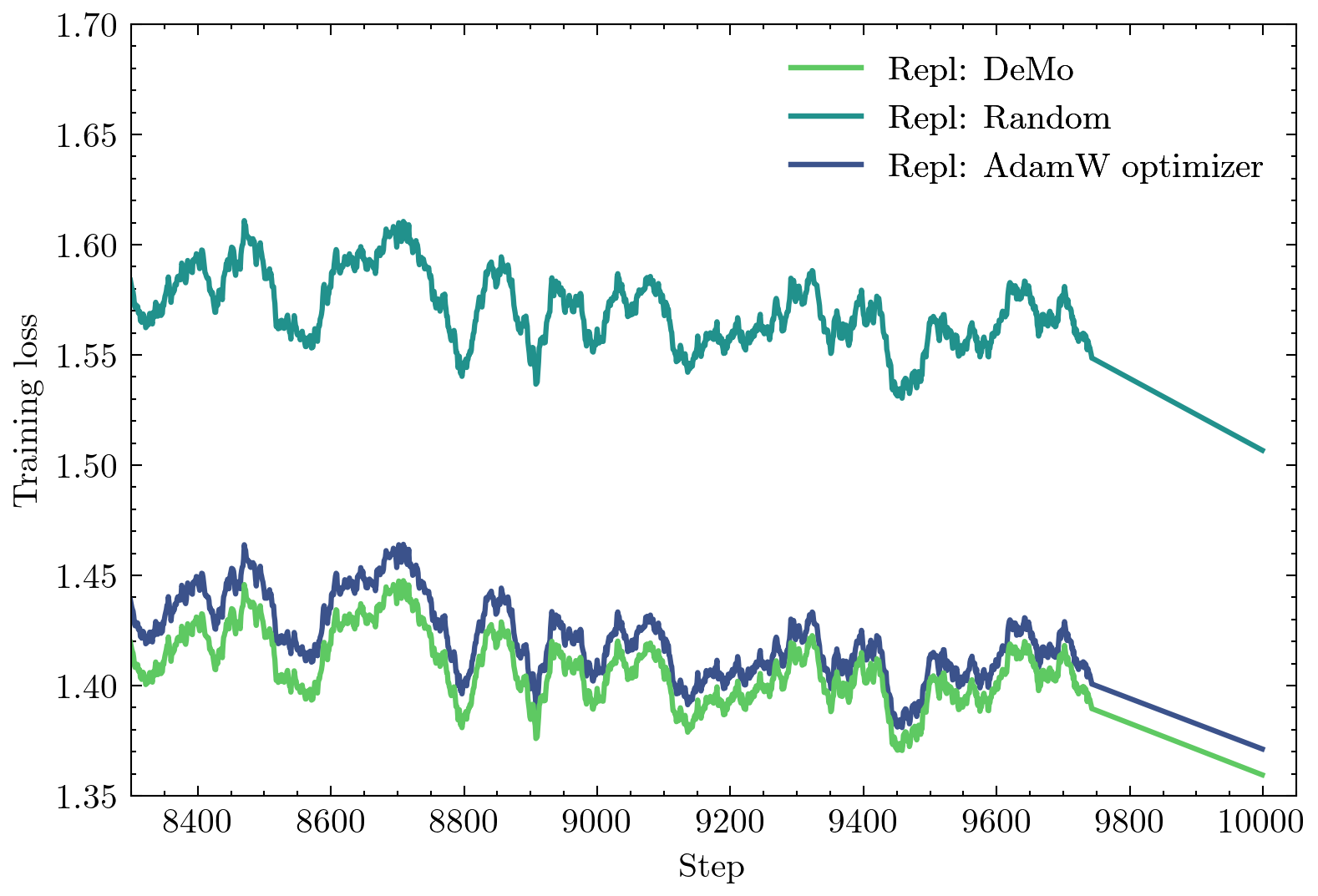}
	\caption{OLMo2 1B train loss (zoomed) over 10K training steps on Dolma v1.6 using DeMo-SGD with a 1/32 compression rate on 64 nodes. We compare replication schemes DeMo and Random against the Hybrid-FSDP + AdamW baseline.}
	\label{fig:olmo:train-loss:scale}
\end{figure}

\begin{figure}[htb]
	\centering
	\includegraphics[width=0.95\linewidth]{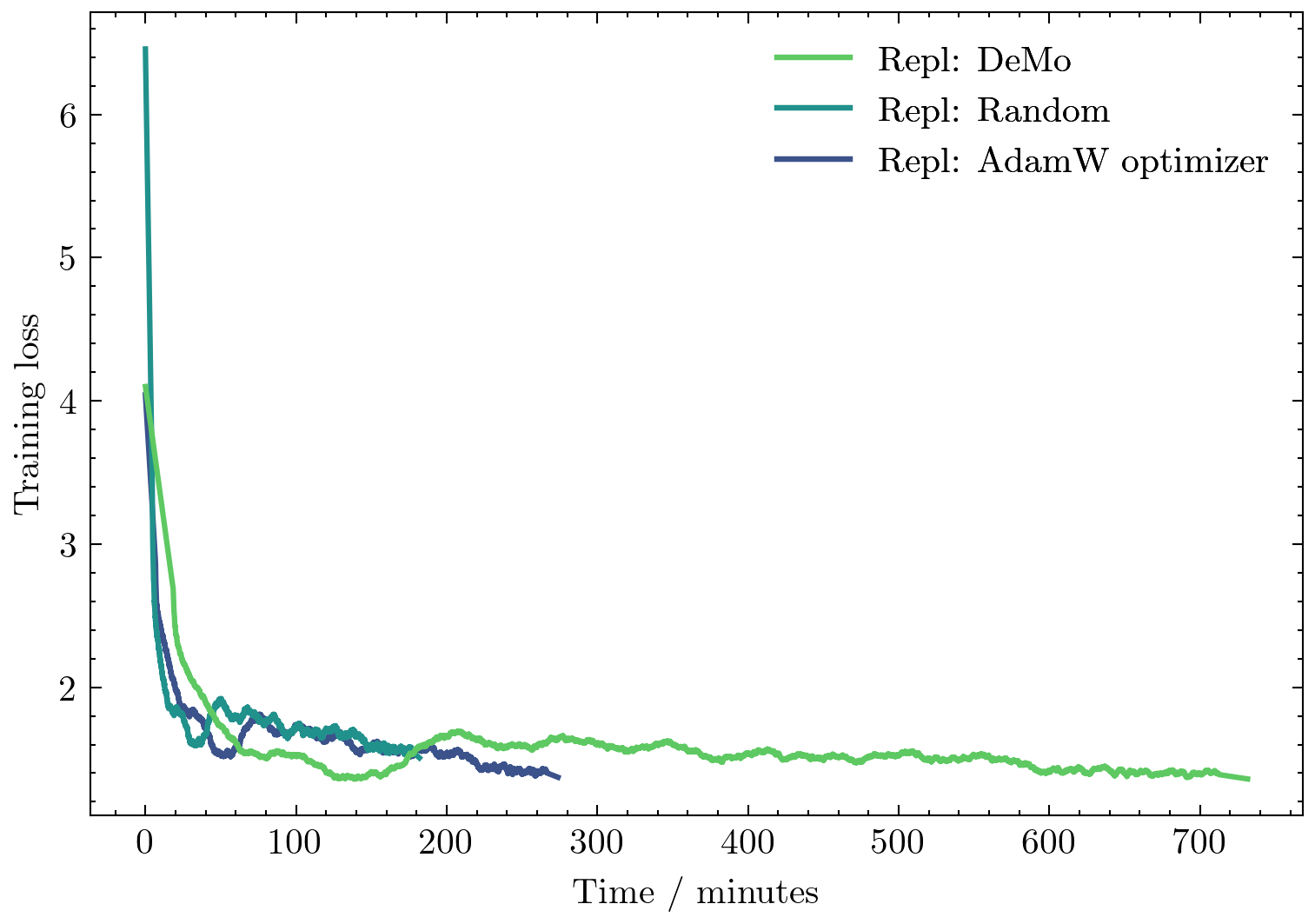}
	\caption{OLMo2 1B train loss vs. wall-clock time (in minutes) on 64 nodes over 10K training steps using DeMo-SGD.
		DeMo does not scale efficiently with node count due to overhead from the \texttt{all\_gather} operation, especially compared to Hybrid-FSDP with AdamW.
	}
	\label{fig:random:train-loss-time:scale}
\end{figure}

\paragraph{Results} We report our findings in Figure \ref{fig:olmo:train-loss:scale}. We see that Random replication performs worse than DeMo replication, which performs almost comparable to conventional AdamW (full-sync). However, DeMo is substantially slower. We attribute this to DeMo's dependence on the \texttt{all\_gather} operation, which prevents it to scale as favorable as the Random replicator. This is shown in Figure \ref{fig:random:train-loss-time:scale}. 

\paragraph{Discussion}
In Figures \ref{fig:olmo:train-loss:scale} and \ref{fig:random:train-loss-time:scale}, we investigate the scale both Random and DeMo replication schemes, powered by the DeMo-SGD optimizer and comparing against conventional Hybrid-FSDP with AdamW optimization. While DeMo replications performance is clearly best, there is a big drawback in the time-scaling of the scheme, due to the \texttt{all\_gather} operation scaling badly, as one would expect, due to the blocking nature of the operation. However, the Random replicator is around 64\% faster than the conventional setup. This could call for 2-stage distributed training setup, employing Random replication for the majority of the training, using the conventional setup for a subsequent stage. 

\section{General Discussion }

Through a set of experiments, we have validated important hyper-parameters, investigating the characteristics of FlexDemo, with DeMo replication, including communication data type, underlying optimizer, and choice of the \texttt{sign} function and Top$K$. We do expect parameters to be architecture and potentially domain-dependent.

We demonstrate that using \texttt{sign} before synchronizing is a corner-stone in this optimization scheme. Immediately, this tells us that direction is more important than magnitude for our case of decoupled optimization. This has also been studied in prior work \cite{bernstein2018signsgd}. The ternary system enables compressing the data even more.

In experiments with full control over the network bandwidth between the nodes, we have shown how Random replication scales better across different replicators and compression rates. This, with the DeMo-SGD optimizer, is not surprising, as SGD employs fewer parameters per model parameter than AdamW.

We have introduced a decoupled variant of AdamW, investigating a version not sharing the first and second moments. However, we demonstrate in Figure \ref{fig:optimchoice:valid} of the Appendix, that Decoupled-AdamW is not generally superior and only outperforms DeMo-SGD when compared on the DiLoCo replicator. We did not develop this optimizer further, as the OLMo2 experiment has shown superior performance of Hybrid-FSDP with AdamW.

We have studied the Striding and Random replication schemes we introduced to facilitate FlexDeMo and challenged experimental choices made in DeMo. Using any of these two schemes enables us to utilize more of the bandwidth sending actual data and not the corresponding indices. We demonstrate that DeMo works best on the ViT and decoder-architectures, whereas Random performs best in the encoder-decoders architecture.

Throughout our experiments, we have experimented with both NCCL and RCCL communications-backends, demonstrating DeToNATION and FlexDeMo across both platforms. 
Overall, this opens an avenue of future research of decoupled training-techniques, extending these schemes but also, importantly, investigating sparse optimization in different domains. This is important, as we demonstrated that different replicators are optimal for different tasks. Future work may study whether the ideal type of replicator depends on the task or architecture, potentially warranting the development of architecture-optimized replicators.

\paragraph{Future Work}

Introducing FSDP requires an initial \texttt{reduce\_scatter} operation on each gradient. PyTorch's FSDP implementation performs this during the backward pass, which we disabled with \texttt{no\_sync}. This can be implemented using hooks to enable overlap with the backward pass. Additionally, asynchronous communication with CUDA streams contains substantial efficiency potential. Adopting FSDP2 or SimpleFSDP~\cite{zhang2024simplefsdp} will further accelerate basic communication operations. These improvements will enable cross-node sharding for very large models and facilitate connecting multiple HPC systems for collaborative training.

\section{Conclusion}

We have shown that FlexDeMo successfully extends DeMo optimization to hybrid sharded FSDP, achieving comparable or better results than AdamW while being significantly faster. DeMo-SGD has displayed superior performance as the underlying optimizer in most settings, while replicator choice depends on architecture and task. DeMo replication demonstrates the best convergence but scales poorly with node count due to \texttt{all\_gather} overhead, whereas Random replication maintains efficiency at scale despite slightly worse loss. FlexDeMo enables training LLMs that exceed single-accelerator memory and supports low-bandwidth network settings, lowering barriers for practitioners and researchers while improving compute efficiency and reducing environmental impact.

\section*{Acknowledgments} 
We acknowledge EuroHPC JU for awarding the projects EUHPC\_A04\_086 access to Leonardo BOOSTER and DeiC-SDU-S5-202500013 for access to LUMI and Ordbogen A/S for providing resources for bandwidth constrained experiments. This research was in parts supported by the Danish Foundation Models (DFM) project.

\appendix

\section{Reproducibility}
All code used to run the experiments in the paper, is made available on GitHub\footnote{https://github.com/schneiderkamplab/DeToNATION/}. The \texttt{\textbackslash benchmarks\textbackslash} folder contain folders for ViT, OLMo2, and T5, each with a \texttt{README.md} file, a \texttt{requirements.txt} file, and a \texttt{train.py} file to start the training. To reproduce the experiments, simply install the packages from the requirements file, and run the \texttt{train.py} file. A more comprehensive guide is available in each of the \texttt{README.md} files. 

Note that some HPCs are restricted in terms of downloads from the web on the compute nodes, and the models and datasets should thus be download accordingly to each setup. Scripts are included to download the datasets, where relevant.

\bibliography{main}

\section{Reproducibility Checklist}
\begin{itemize}
	\item Includes a conceptual outline and/or pseudocode description of AI methods introduced: \textbf{Yes}
	\item Clearly delineates statements that are opinions, hypothesis, and speculation from objective facts and results: \textbf{Yes}
	\item Provides well marked pedagogical references for less-familiare readers to gain background necessary to replicate the paper: \textbf{Yes}
	\item Does this paper make theoretical contributions? \textbf{No} 
	\item Does this paper rely on one or more datasets?: \textbf{Yes}
	\begin{itemize}
		\item A motivation is given for why the experiments are conducted on the selected datasets \textbf{Yes}
		\item All novel datasets introduced in this paper are included in a data appendix. \textbf{No}
		\item All novel datasets introduced in this paper will be made publicly available upon publication of the paper with a license that allows free usage for research purposes. \textbf{Yes}
		\item All datasets drawn from the existing literature (potentially including authors’ own previously published work) are accompanied by appropriate citations \textbf{Yes}. 
		\item All datasets drawn from the existing literature (potentially including authors’ own previously published work) are publicly available. \textbf{Yes}
		\item All datasets that are not publicly available are described in detail, with explanation why publicly available alternatives are not scientifically satisficing. \textbf{NA}.
	\end{itemize}
	\item Does this paper include computational experiments? \textbf{Yes}.
	\begin{itemize}
		\item This paper states the number and range of values tried per (hyper-) parameter during development of the paper, along with the criterion used for selecting the final parameter setting \textbf{Yes}
		\item Any code required for pre-processing data is included in the appendix. \textbf{Yes}.
		\item All source code required for conducting and analyzing the experiments is included in a code appendix.\textbf{Yes}
		\item All source code required for conducting and analyzing the experiments will be made publicly available upon publication of the paper with a license that allows free usage for research purposes \textbf{Yes}
		\item All source code implementing new methods have comments detailing the implementation, with references to the paper where each step comes from \textbf{No}
		\item If an algorithm depends on randomness, then the method used for setting seeds is described in a way sufficient to allow replication of results. \textbf{Yes}
		\item This paper specifies the computing infrastructure used for running experiments (hardware and software), including GPU/CPU models; amount of memory; operating system; names and versions of relevant software libraries and frameworks \textbf{Partial}
		\item This paper formally describes evaluation metrics used and explains the motivation for choosing these metrics. \textbf{No} 
		\item This paper states the number of algorithm runs used to compute each reported result. \textbf{Yes}
		\item Analysis of experiments goes beyond single-dimensional summaries of performance (e.g., average; median) to include measures of variation, confidence, or other distributional information. \textbf{No}
		\item The significance of any improvement or decrease in performance is judged using appropriate statistical tests (e.g., Wilcoxon signed-rank). \textbf{No}
		\item This paper lists all final (hyper-)parameters used for each model/algorithm in the paper’s experiments \textbf{Yes}
	\end{itemize}
\end{itemize}

\clearpage

\section{A: Communication DeMo vs. FlexDeMo}
\begin{figure*}
	\centering
	\begin{subfigure}[b]{0.31\textwidth}
			\centering
			\includegraphics[width=\textwidth]{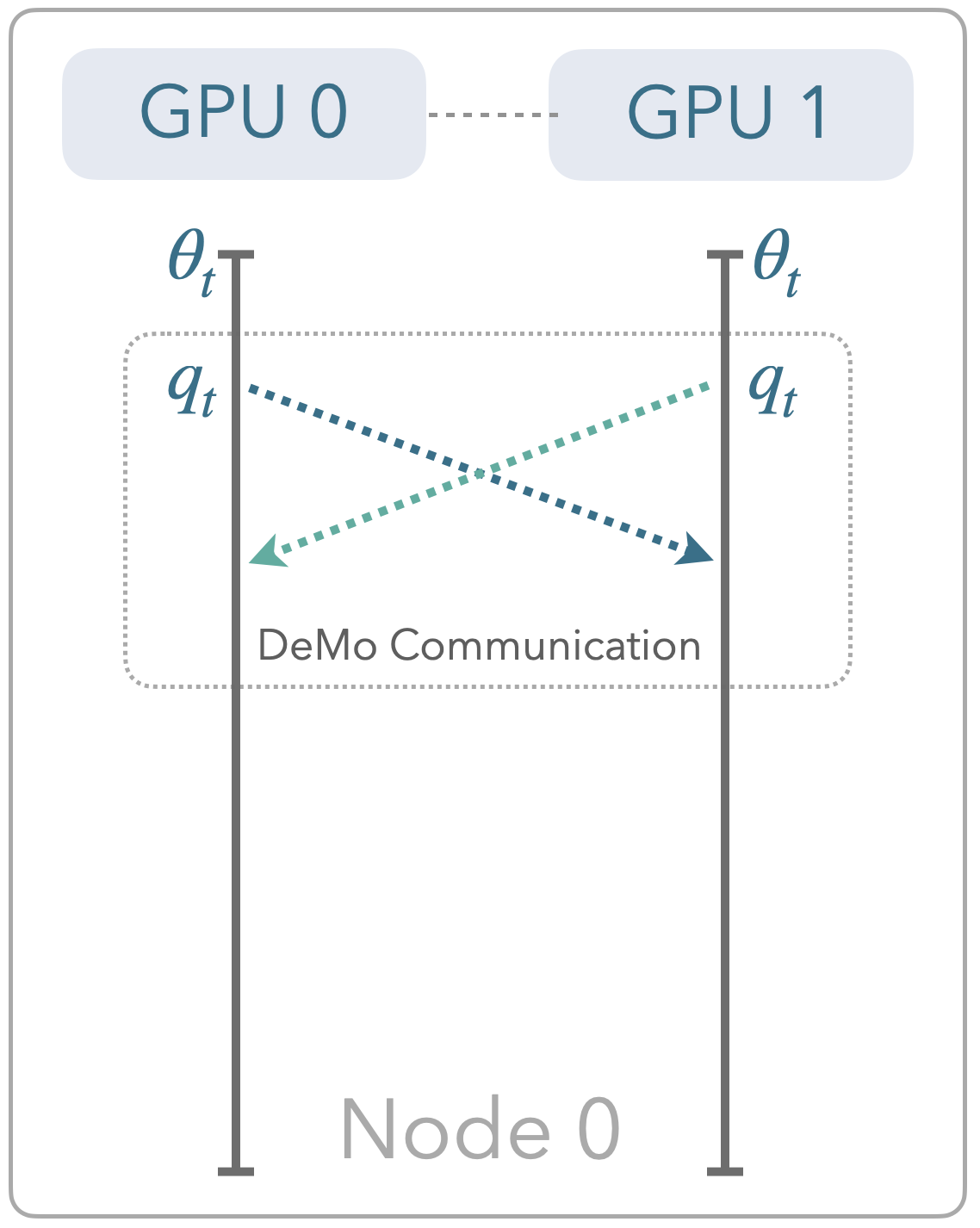}
			\caption{DeMo - 1 Node}
			\label{fig:DeMo_communication_pattern_1_node}
		\end{subfigure}
	\hfill
	\begin{subfigure}[b]{0.64\textwidth}
			\centering
			\includegraphics[width=\textwidth]{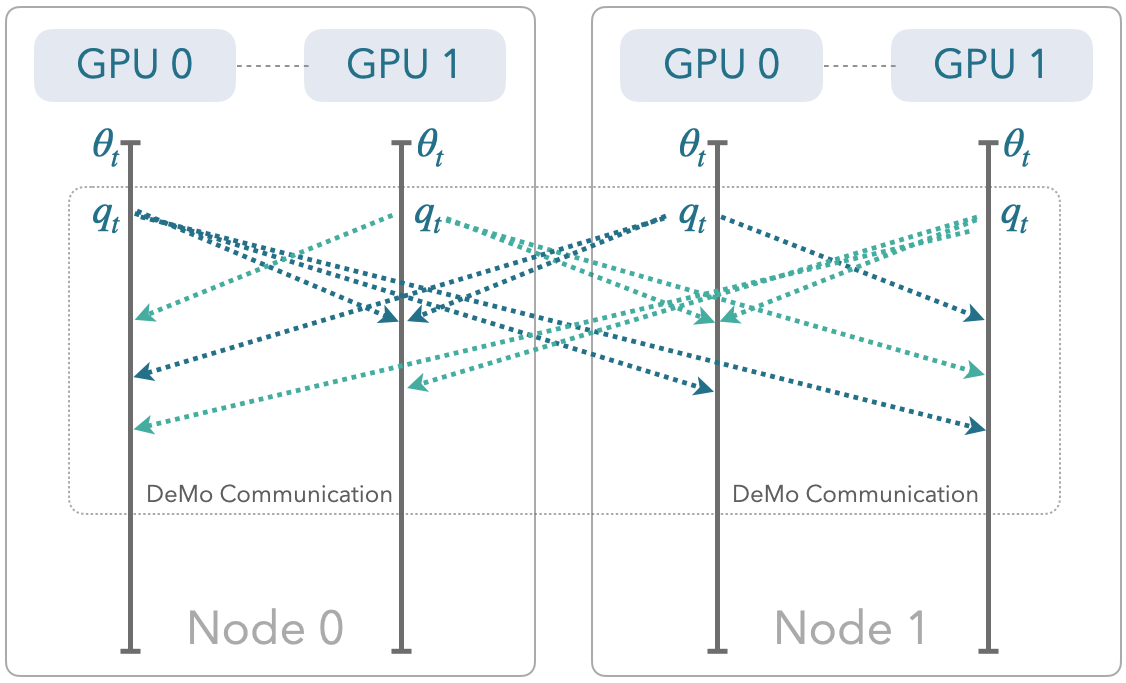}
			\caption{DeMo - 2 Nodes}
			\label{fig:DeMo_communication_pattern_2_nodes}
		\end{subfigure}
	\vspace{2em}
	\centering
	\begin{subfigure}[b]{0.7\textwidth}
			\centering
			\includegraphics[width=\textwidth]{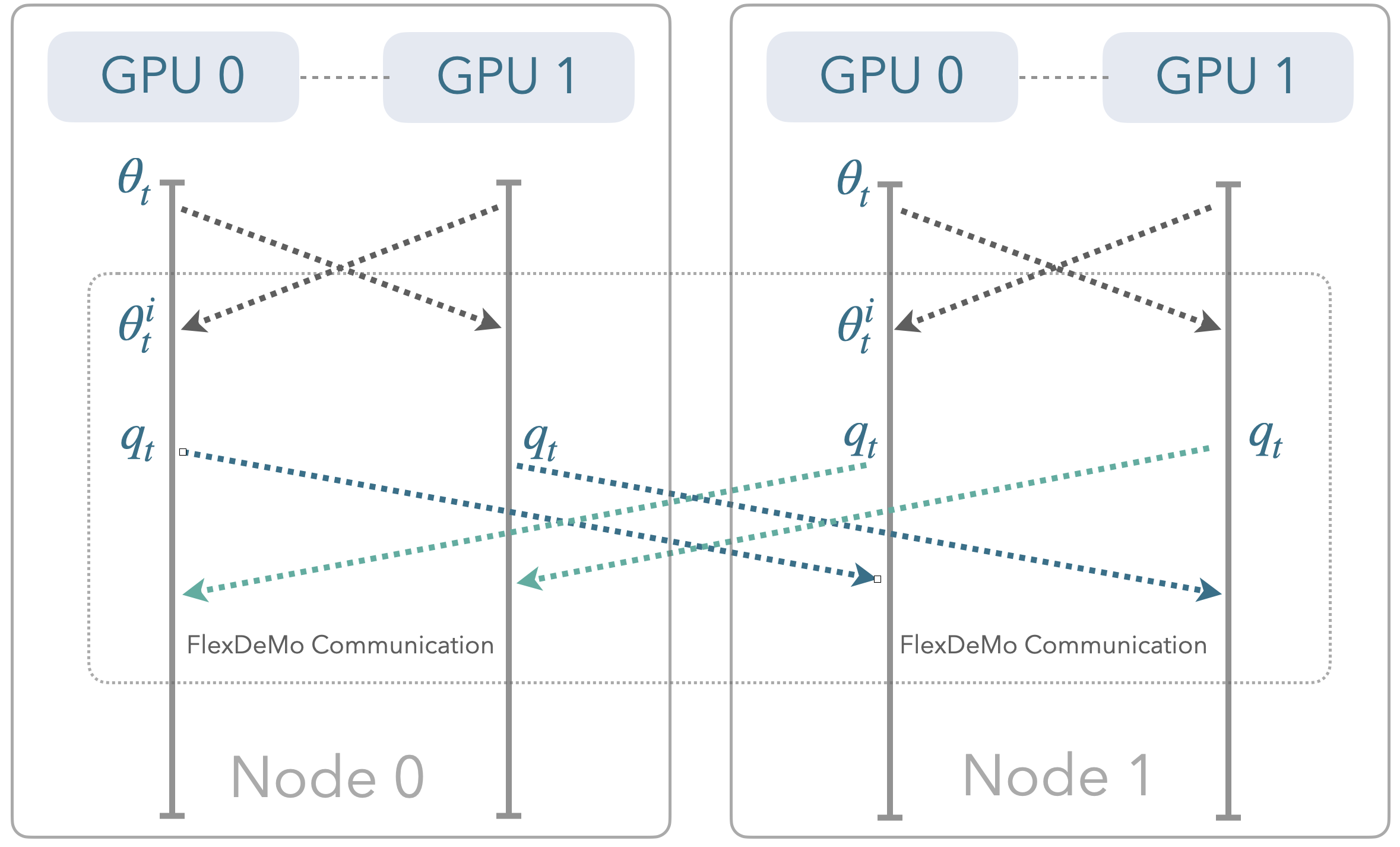}
			\caption{FlexDeMo - 2 Nodes}
			\label{fig:flexDeMo_communication_pattern}
		\end{subfigure}
	\caption{
			Communication Patterns of DeMo and FlexDeMo, with their respective intra-node and inter-node communication.
		}
	\label{fig:comm_pattern}
\end{figure*}

In Figure \ref{fig:comm_pattern} we give a detailed visualization of the communication in DeMo and FlexDeMo using hybrid-sharding.

\section{B: DeMo Hyperparameters}
Here, we study hyperparameters for DeMo replication, including \texttt{sign}, TopK, transfer-\texttt{dtype}. The experiments are conducted with Hybrid-FSDP on two nodes with two accelerators each. We run training for 20 epochs without weight decay. The optimizer and the replication scheme depends on the specific experiment.

\begin{figure}[H]
	\centering
	\includegraphics[width=\linewidth]{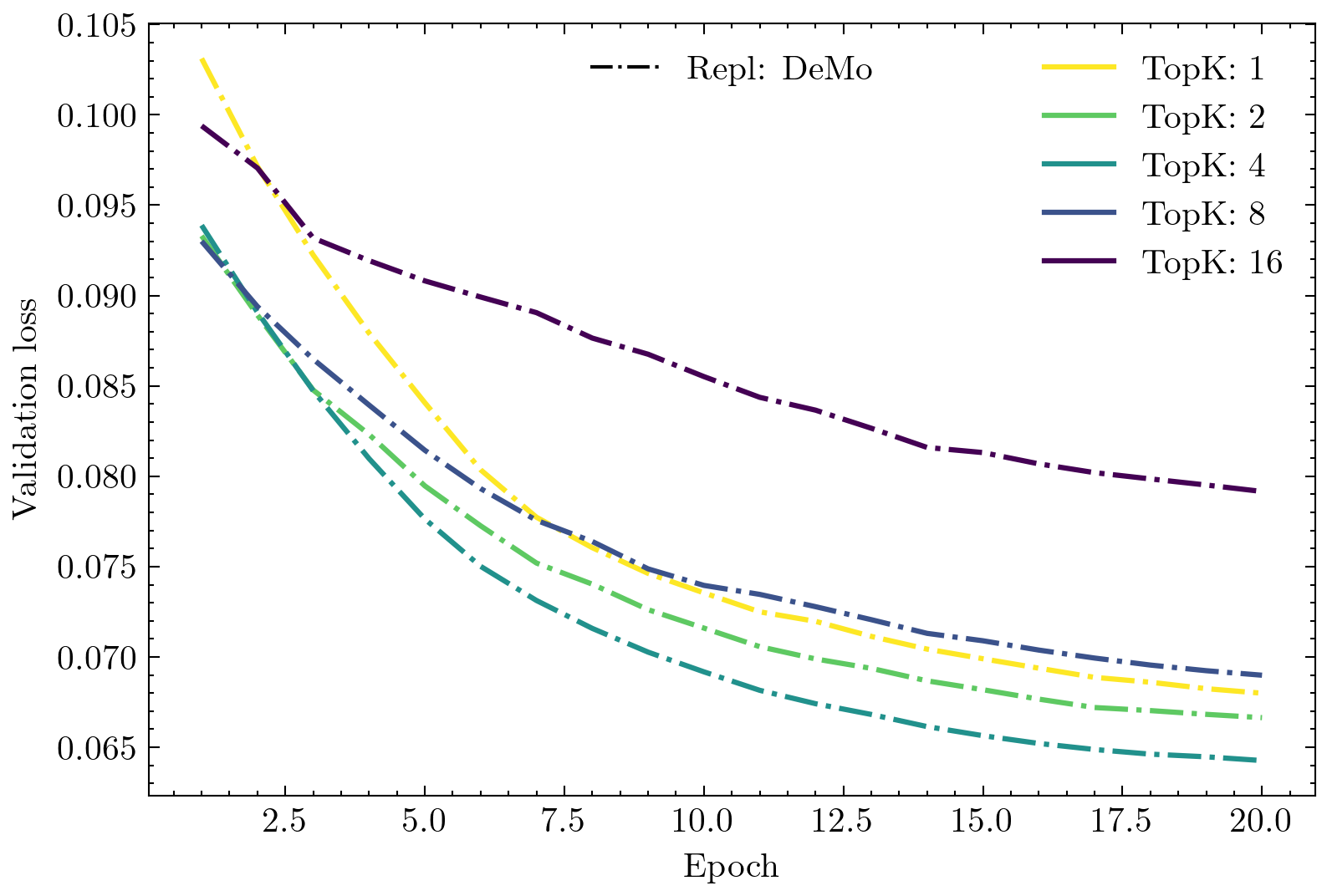}
	\caption{T5-Large on OpusBooks En-Fr over different TopK's. Demo-SGD optimizer and DeMo-replicator.}
	\label{fig:topk:valid}
\end{figure}

\paragraph{TopK}\label{appendix:topk}
\texttt{TopK} hyperparameter, defines the number of fastest moving momenta which is synchronized between replications, when using the DeMo replicator. We conduct an experiment, investigating the optimal TopK, for our setup. We report our results in Figure \ref{fig:topk:valid}. Top$4$ is clearly superior for our setting, followed by Top$2$, Top$1$, Top$8$, respectively. Last, with a huge degradation in performance Top$16$.

\begin{figure}[H]
	\centering
	\includegraphics[width=\linewidth]{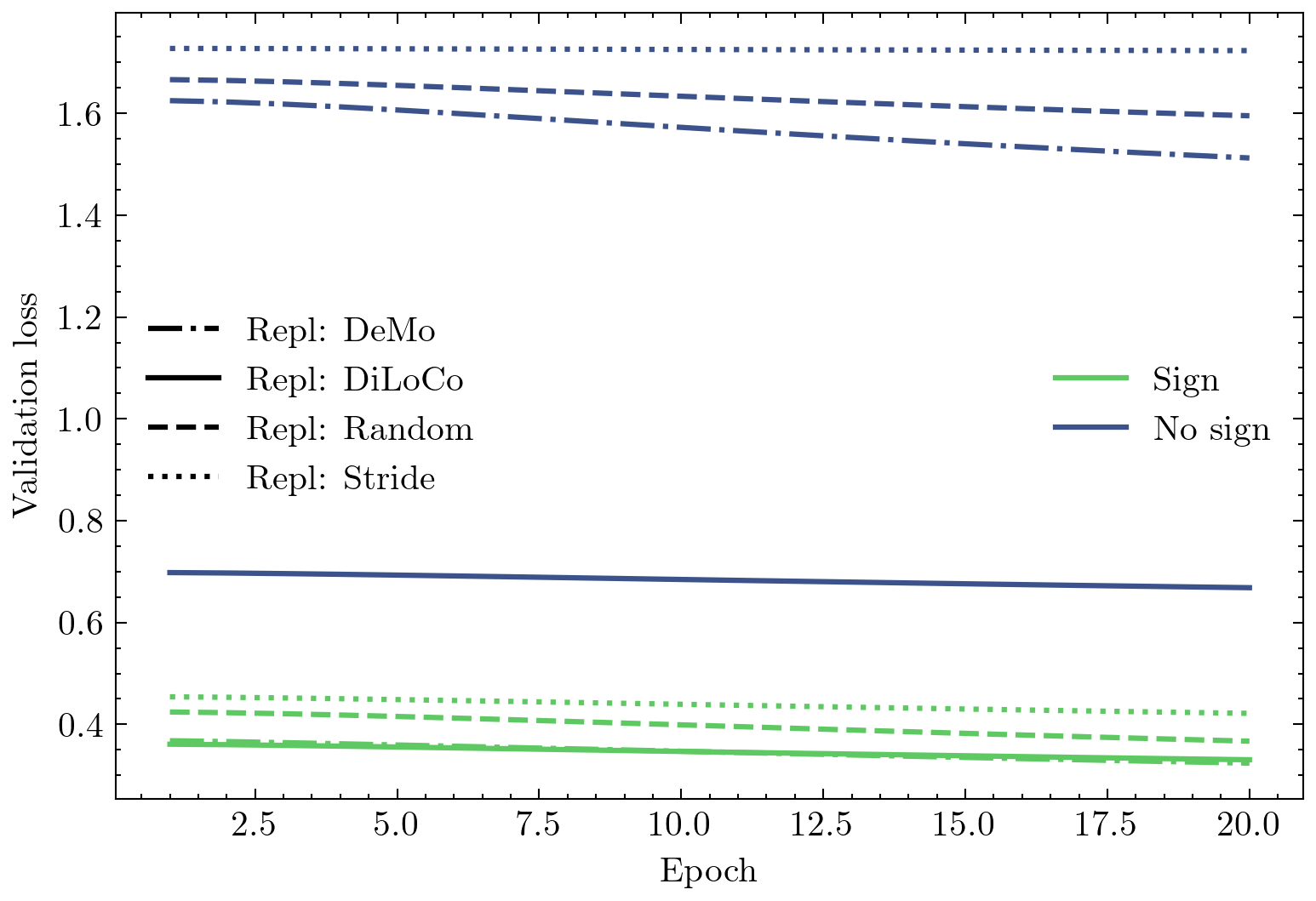}
	\caption{Validation loss evaluations on T5-Large trained on OpusBooks (En-Fr) demonstrate the impact of employing the sign function prior to gradient synchronization. Very clearly \texttt{sign} have a general positive impact, especially for the sparse optimization schemes.}
	\label{fig:signnosign:validation}
\end{figure}

\paragraph{Sign vs. No Sign.}\label{signnosign}
The \texttt{sign}-function\footnote{https://docs.pytorch.org/docs/2.7/generated/torch.sign.html} introduces great reduction in required data transferred, reducing the actual communication to a ternary system with minimal information exchange. This design-choice was introduced in DeMo \citep{peng2024demodecoupledmomentumoptimization}, however, not justified. We report loss on the validation set in Figure \ref{fig:signnosign:validation}. Interestingly, employing the \texttt{sign}-function to gradients before synchronization generally yields substantial performance gains with a large margin in the validation results, with DeMo and DiLoCo yielding the strongest performance, closely followed by Random, while Striding is underperforming.

\begin{figure}
	\centering
	\includegraphics[width=\linewidth]{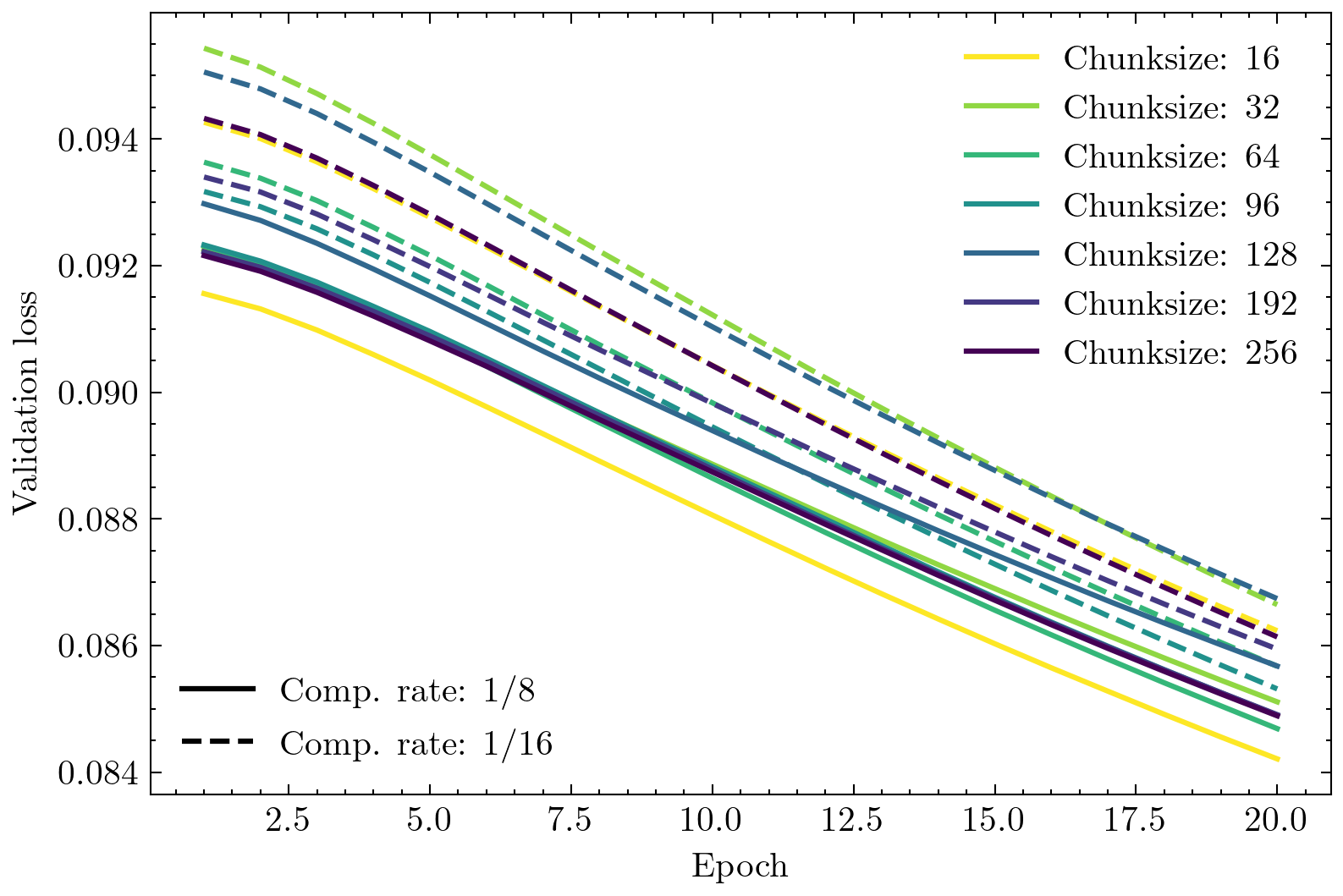}
	\caption{SGD DeMo Validation Loss Curve for Compression Rates $1/16$ and $1/8$ over the chunk-sizes $16,32,64,96,18,192$ and $256$.}
	\label{fig:demo:chunksizes}
\end{figure}

\begin{figure}
	\centering
	\includegraphics[width=\linewidth]{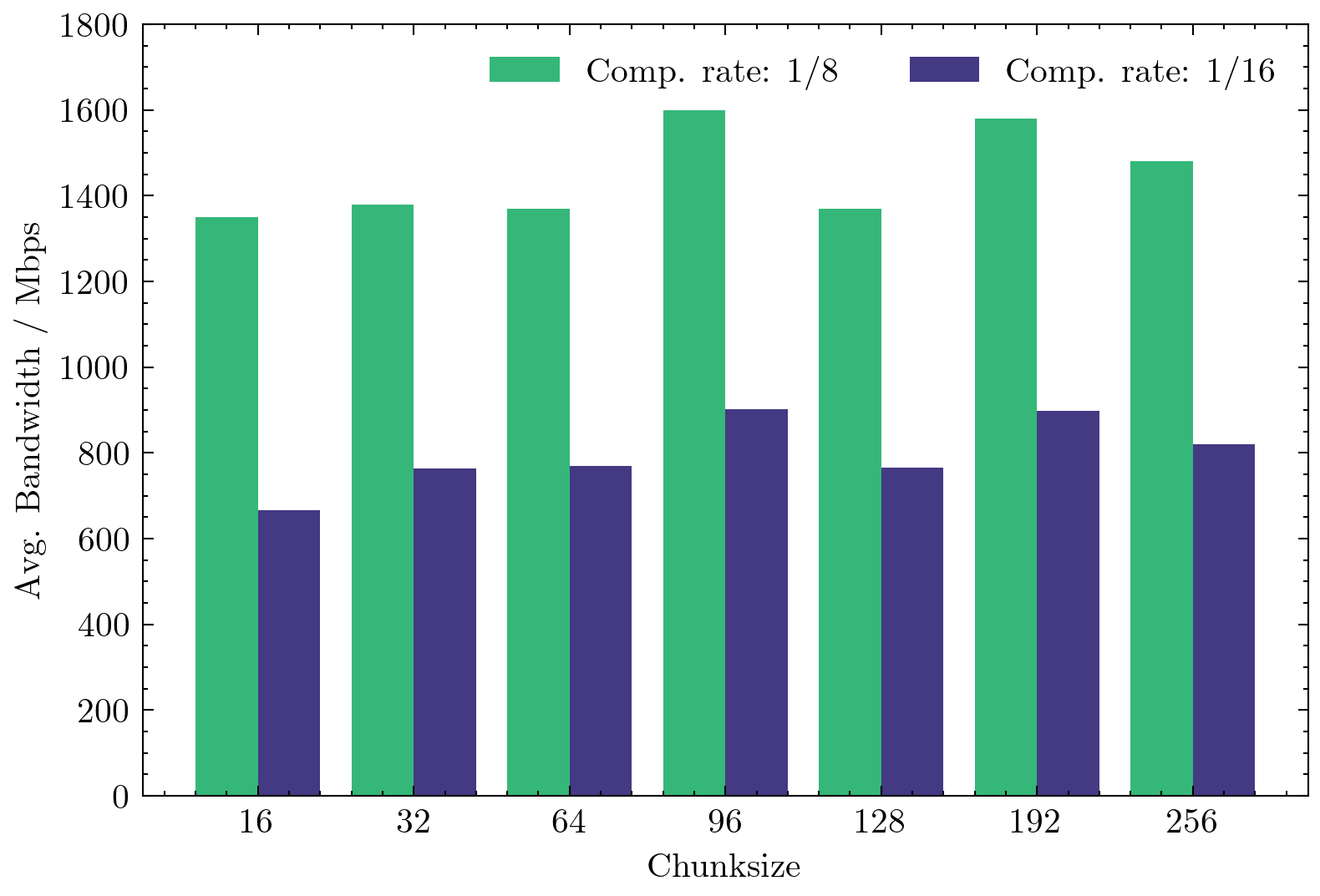}
	\caption{Bandwidth usage for chunk size variations.}
	\label{fig:demo:chunksize:bw}
\end{figure}

\paragraph{DeMo Chunk Size}\label{appendix:chunksize}
We conduct experiments with the DeMo-replicator across different chunk-sizes with compression rates of $1/16$ and $1/8$, investigating the training behavior and performance yields. We report our findings in Figure \ref{fig:demo:chunksizes}, with the corresponding bandwidth usage in Figure \ref{fig:demo:chunksize:bw}. A compression rate of $1/8$ and the smallest chunk-size of 16 demonstrates slightly better performance, generally getting worse as the chunk-size increases. Interestingly, we see the opposite tendency for a compression rate $1/16$, where the larger chunk sizes yields better performances. Compression rate $1/16$ with chunk-size 96 outperforms chunk-size $128$ with compression-rate $1/8$. However, this is not a general tendency.

\begin{figure}
	\centering
	\includegraphics[width=\linewidth]{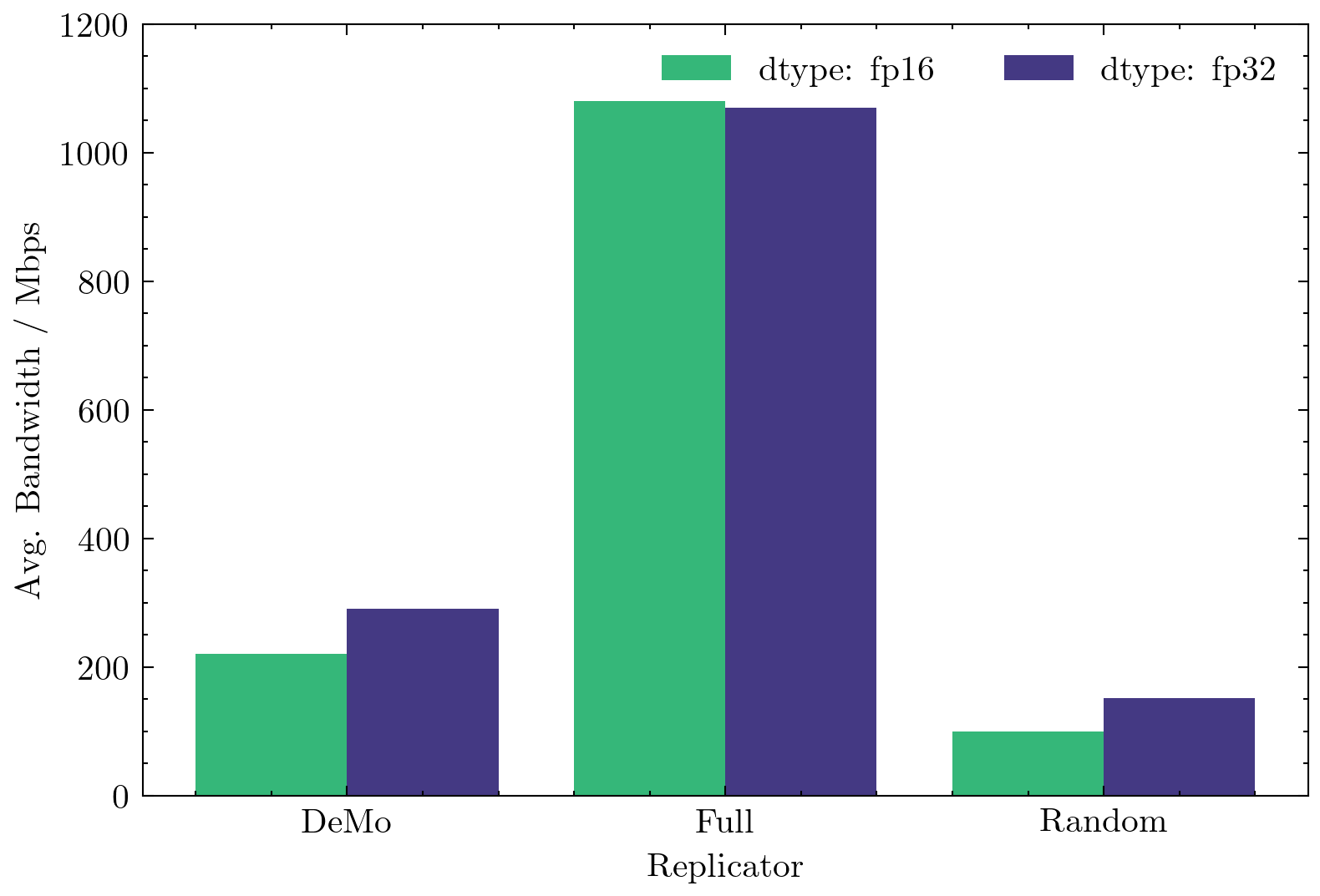}
	\caption{Bandwidth usage for each precision-type.}
	\label{fig:comm:bw}
\end{figure}

\begin{figure}
	\centering
	\includegraphics[width=0.95\linewidth]{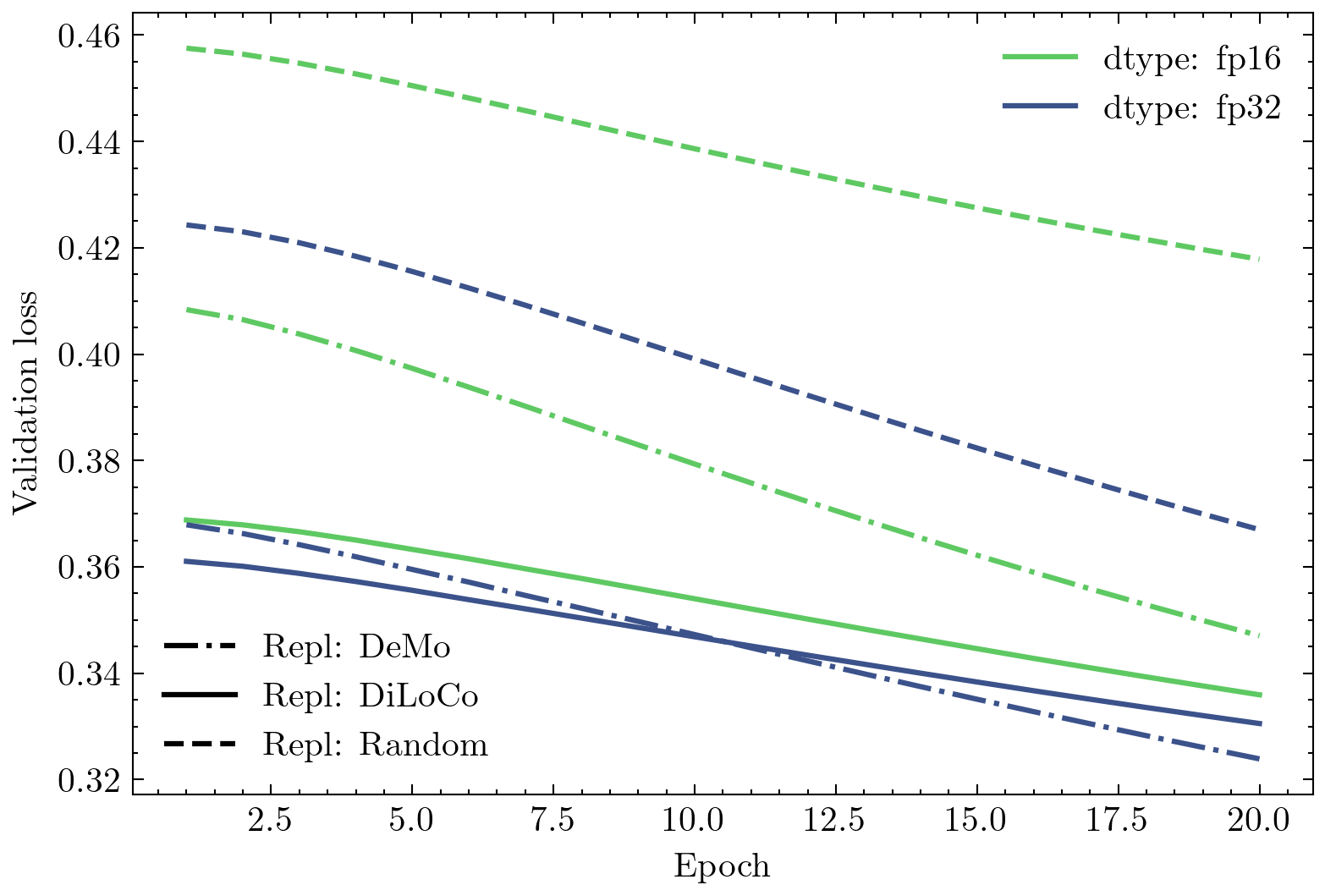}
	\caption{Validation loss for different precision data types.}
	\label{fig:comm:dtype}
\end{figure}

\paragraph{Communication Floating Point Data Type (dtype)}\label{appendix:comdtype}
The most significant momenta are shared in full precision by previous work \citep{peng2024demodecoupledmomentumoptimization}. Naturally, we need to confirm the necessity to transfer double precision on the cost of doubled data footprint, visualized in Figure \ref{fig:comm:bw}. We validate this choice in Figure \ref{fig:comm:dtype}, where we confirm DeMo and full precision is superior, followed close by full sync (DiLoCo) full-precision. Interestingly, the full-sync scheme is mostly unaffected by the data precision, whereas both DeMo and Random replication change significantly with precision, where higher precision, leads to better performance.

\section{C: Experiment Train Losses}\label{appendix:parametersearch:t5_trainloss}\label{appendix:parametersearch:vit_trainloss}
\begin{figure}
	\centering
	\begin{subfigure}[b]{\linewidth}
			\centering
			\includegraphics[width=\linewidth]{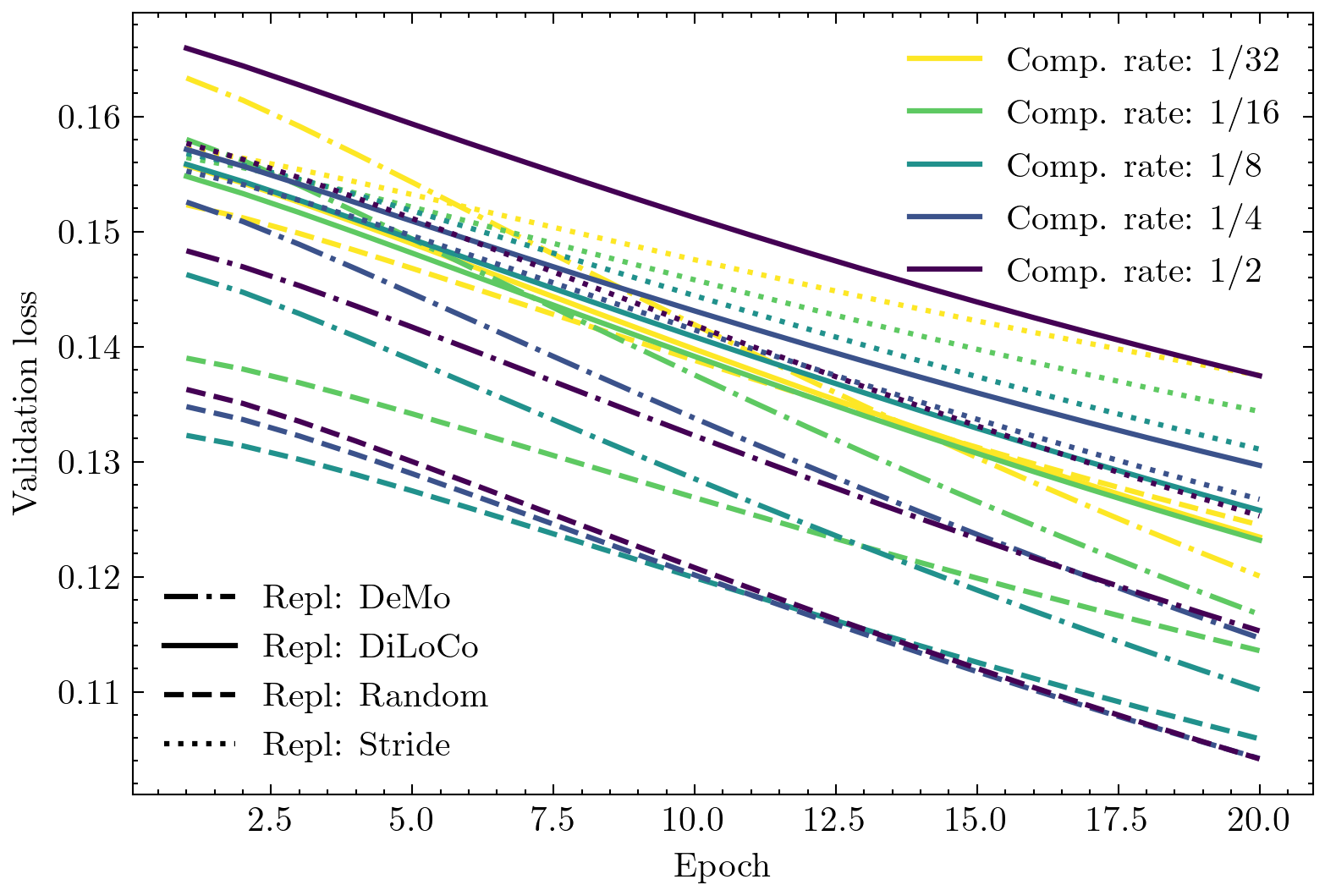}
		\end{subfigure}
	\caption{T5-Large Train loss on the Opus Books En-Fr subset.}
	\label{appendix:parametersearch:trainloss:t5}
\end{figure}

\begin{figure}
	\centering
	\begin{subfigure}[b]{\linewidth}
			\centering
			\includegraphics[width=\linewidth]{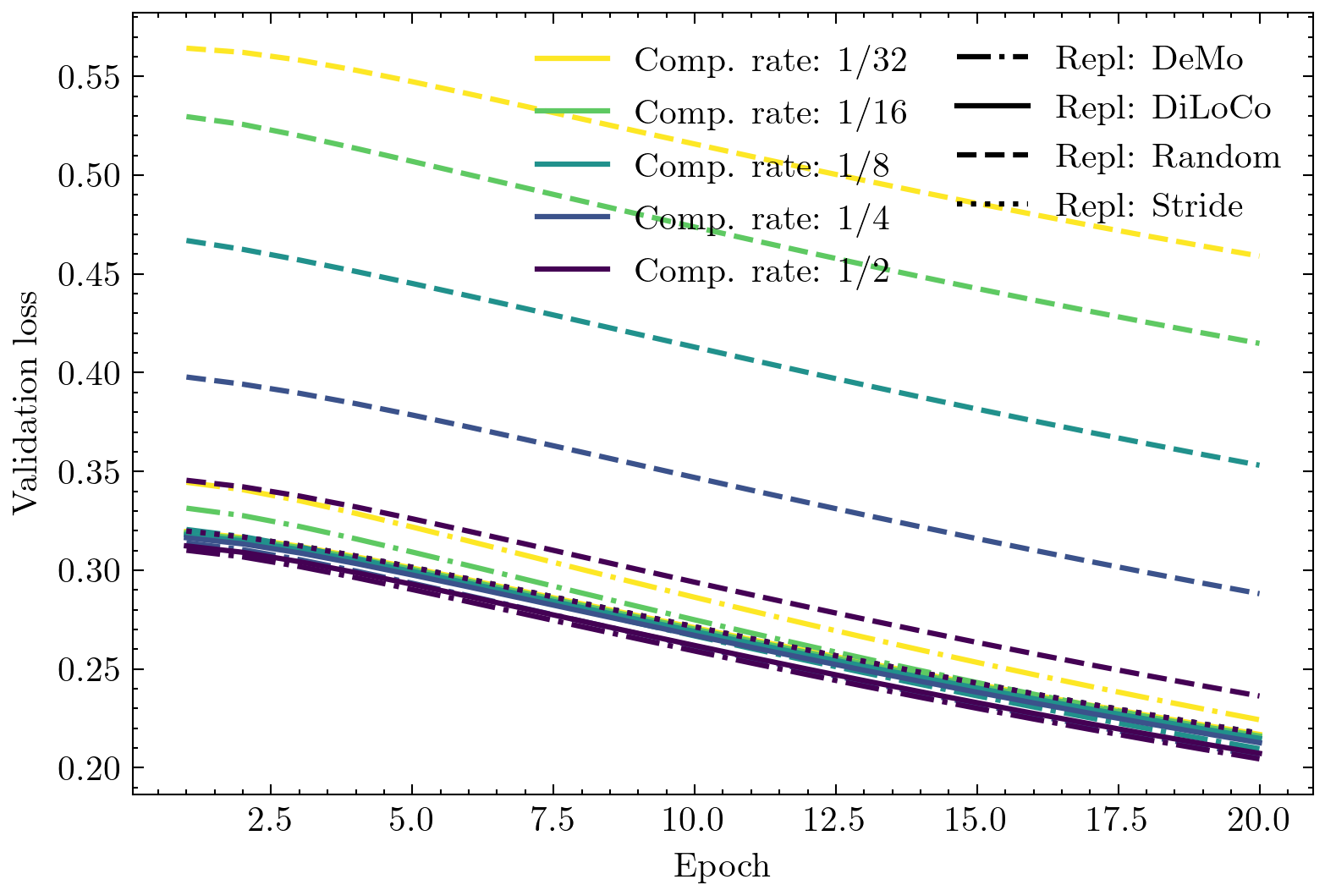}
		\end{subfigure}
	\caption{Epoch train loss for Vit-B 224x224 on Cifar100}
	\label{appendix:parametersearch:trainloss:vitb}
\end{figure}

\begin{figure}
	\centering
	\includegraphics[width=0.95\linewidth]{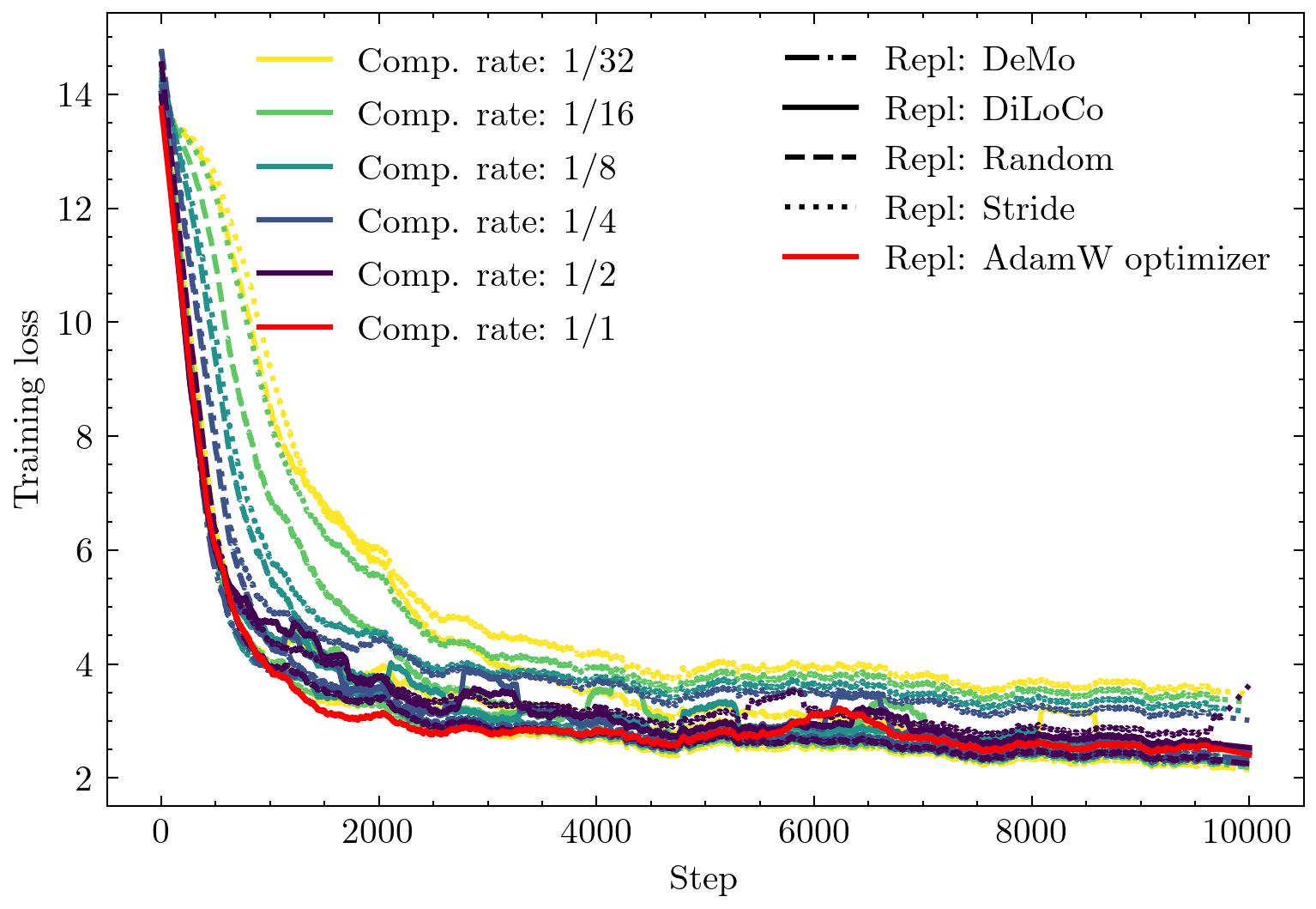}
	\caption{OLMo2 1B train loss over 10K training steps on Dolma v1.6 using different replicators and compression rates. All experiments, except the Hybrid-FSDP baseline with AdamW on 2 nodes, use DeMo-SGD.}
	\label{fig:olmo:train-loss-nonzoom}
\end{figure}

Here, we report train-loss curves for the experiments in the main paper. In Figure \ref{appendix:parametersearch:trainloss:t5} we report the train-losses for the T5 translation task experiments, and in Figure \ref{appendix:parametersearch:trainloss:vitb} we report the train-losses for the ViT image classification experiments. Figure \ref{fig:olmo:train-loss-nonzoom} shows the non-zoomed training loss of the OLMo2 experiment in the main paper (Figure \ref{fig:olmo:train-loss}.

\section{D: Average Time per Optimizer Step}
\begin{figure}
	\centering
	\includegraphics[width=\linewidth]{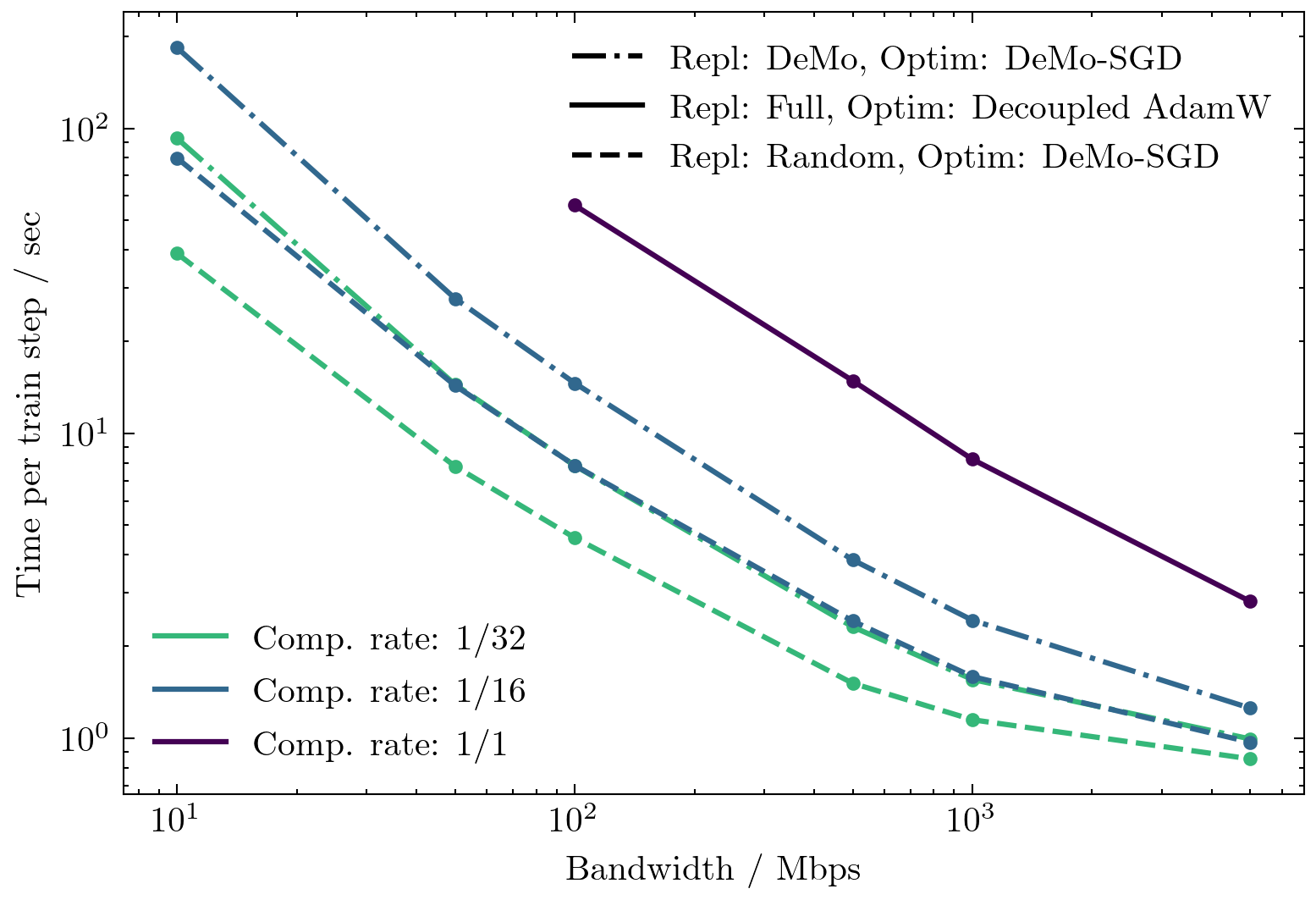}
	\caption{Average time per optimizer step for T5-Large on 2 nodes across employing different replicators using DeMo-SGD and Decoupled AdamW as base optimizers}
	\label{fig:avgstepspeed:t5}
\end{figure}

Figure \ref{fig:avgstepspeed:t5} shows the average time per optimizer step for T5-Large on 2 nodes across different replicators and bandwidth limits.

\section{E: Ethical Considerations}
Our work improves efficiency of distributed training. It may contribute to the democratization of training large language models and to reducing the environmental footprint of LLMs. In particular, our findings suggest that LLMs potentially can be trained on a lower budget. We acknowledge that this may make training such models more accessible. However, this work is purely scientific and does not promote easier access to systems for training such models. Using our proposed method, it might be possible to reduce compute budgets, making larger models more financially accessible. 

\section{F: SGD vs. Decoupled AdamW}\label{sgdvsadamw}
\begin{figure}[htb]
	\centering
	\includegraphics[width=0.95\linewidth]{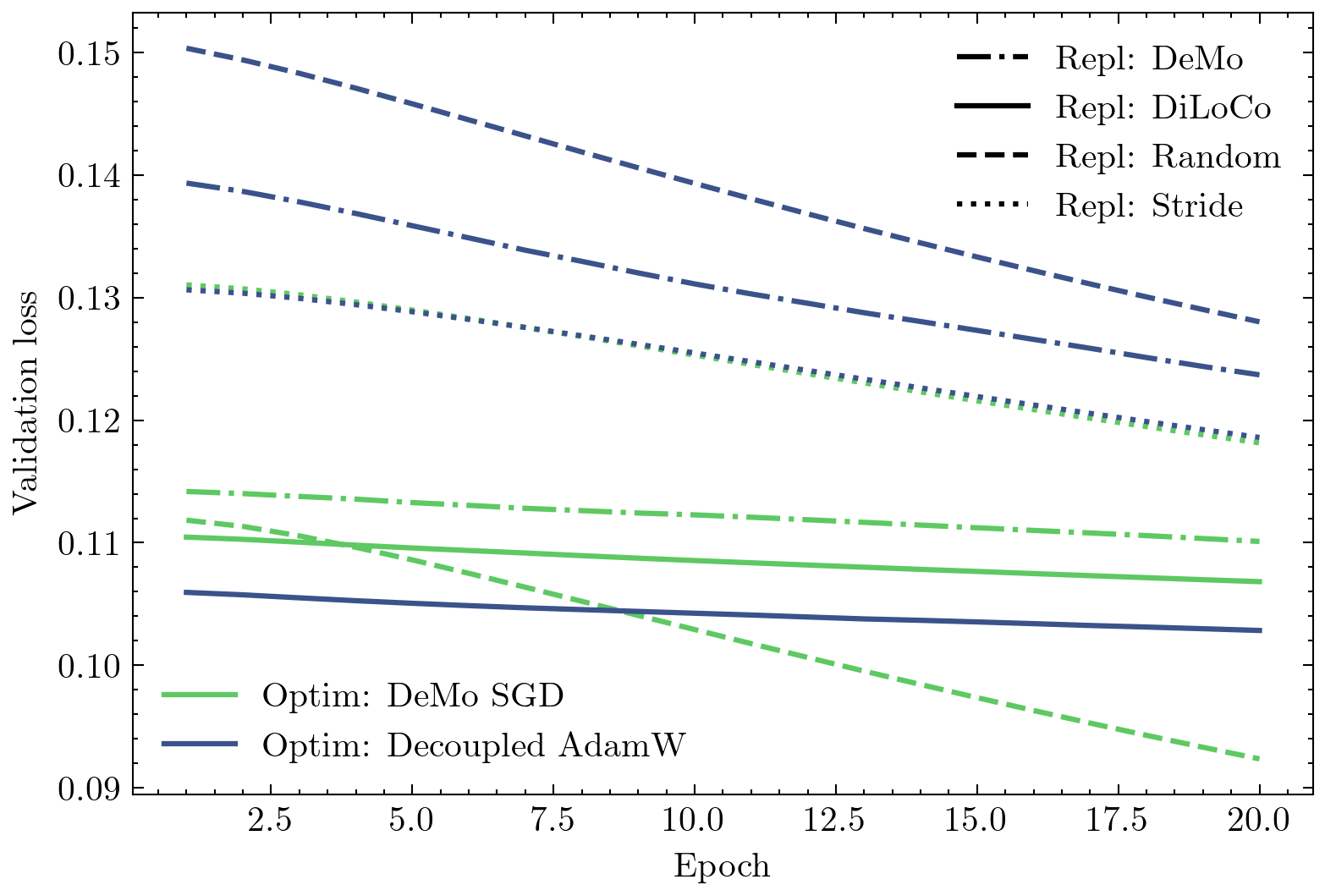}
	\caption{T5-Large OpusBooks English-French, validation loss for DeMo SGD and Decoupled AdamW. The compression rate is chosen to keep the bandwidth usage constant across replication methods. DeMo-SGD with Random replication is clearly superior converging faster than both DiLoCo variants and DeMo SGD with DeMo replicator. Decoupled AdamW behaves similarly to SGD-DeMo with DiLoCo replicator, with an offset on the y-axis}
	\label{fig:optimchoice:valid}
\end{figure}

AdamW \cite{kingma2014adam} is de-facto default optimizer, having demonstrated superior performance across many domains and tasks. However DeMo \citep{peng2024demodecoupledmomentumoptimization} is employing SGD (which we denote DeMo-SGD, we differentiate as it accumulates momenta.) as underlying optimizer. To investigate this design-choice and broaden the investigation, we set out to compare DeMo-SGD and our Decoupled-AdamW with employing DeMo, DiLoCo~\cite{douillard2023diloco}, Random, and Striding replication schemes. We report validation loss in Figure \ref{fig:optimchoice:valid}. The experiments are conducted such that the bandwidth usage is identical in each setting (we denote this as the compression-rate), as it would be defined by hardware in practice. 
DeMo-SGD with Random replication demonstrates best performance. DeMo-SGD yields clearly better at generalization to the validation-set, with Random replication being best. Decoupled-AdamW and DeMo-SGD with DiLoCo replication following as seconds and third best, respectively. DeMo-SGD is improving overall as fast as the DiLoCo variants, but overall worse. Decoupled-AdamW with DeMo and Random replication schemes yields worst performance.

\end{document}